\documentclass{article}
\usepackage[left=2.5cm,right=2.5cm,top=3cm,bottom=3cm,letterpaper]{geometry}
\providecommand{\keywords}[1]{\textbf{\textit{Keywords ---}} #1}
\usepackage[utf8]{inputenc} 
\usepackage[T1]{fontenc}    
\usepackage{hyperref}       
\usepackage{url}            
\usepackage{booktabs}       
\usepackage{amsfonts}       
\usepackage{nicefrac}       
\usepackage{microtype}      

\usepackage{color}
\usepackage{amsmath}
\usepackage{multirow}
\usepackage{graphicx}
\usepackage{pdflscape}

\usepackage{wrapfig}
\usepackage{lipsum}  

\usepackage{algorithm}
\usepackage{algorithmic}

\usepackage{times}
\usepackage{subfigure}

\usepackage{amsthm}
\usepackage{amssymb}

\usepackage{epstopdf}
\usepackage{enumerate}

\usepackage{array}
\newcolumntype{L}{>{\arraybackslash}m{6.2cm}}
\newcolumntype{M}{>{\centering\arraybackslash}m{1.8cm}}
\newcolumntype{N}{>{\arraybackslash}m{16cm}}

\begin{document}
    %
    \title{Estimating Missing Data in Temporal Data Streams Using Multi-directional Recurrent Neural Networks }
    %
    %
    %
    
    \author{Jinsung~Yoon\thanks{J. Yoon is with the Department of Electrical and Computer Engineering, University of California, Los Angeles, CA, 90095 USA; e-mail: jsyoon0823@ucla.edu.}, William~R.~Zame\thanks{W. R. Zame is with the Departments of Economics and Mathematics, University of California, Los Angeles, CA, 90095 USA; e-mail: zame@econ.ucla.edu.}, and~Mihaela~van der Schaar
        \thanks{M. van der Schaar is with the Department of Engineering Science, University of Oxford, OX1 3PJ UK and the Alan Turing Institute, London, UK; e-mail: mihaela.vanderschaar@oxford-man.ox.ac.uk.}
    }
    
    \date{}
    
    \maketitle
    
    \begin{abstract} 
    Missing data is a ubiquitous problem.  It is especially challenging in  medical settings because many streams of measurements are collected at different -- and often irregular -- times.   Accurate estimation of those missing measurements is critical for many reasons, including  diagnosis, prognosis and treatment. Existing methods address this estimation problem  by interpolating within data streams or imputing across data streams (both of which ignore important information) or ignoring the temporal aspect of the data and imposing strong assumptions about the nature of the data-generating process and/or the pattern of missing data (both of which are especially problematic for medical data). We propose a new approach, based on a novel deep learning architecture that we call a Multi-directional Recurrent Neural Network (M-RNN) that interpolates within data streams and imputes across data streams. We demonstrate the power of our approach by applying it to five real-world medical datasets. We show that it provides dramatically improved estimation of missing measurements in comparison to 11 state-of-the-art benchmarks (including Spline and Cubic Interpolations, MICE, MissForest, matrix completion and several RNN methods); typical improvements in Root Mean Square Error are between 35\% - 50\%. Additional experiments based on the same five datasets demonstrate that the improvements provided by our method are extremely robust.    \end{abstract}
    \small{
    \keywords{Missing Data, Temporal Data Streams, Imputation, Recurrent Neural Nets}
    }

    %
    
    \section{Introduction}
    %
    %
    %
    %
    Missing data/measurements present a ubiquitous problem. The problem is especially challenging in medical settings which present time series containing many streams of measurements that are sampled at different and  irregular times, and is especially important in these settings because accurate estimation of these missing measurements is often critical for  accurate diagnosis, prognosis and treatment, as well as for accurate modeling and statistical analyses.  This paper presents a new method  for estimating missing measurements in time series data, based on a novel deep learning architecture. By comparing our method with current state-of-the-art benchmarks on a variety of real-world medical datasets, we demonstrate that our method is much more accurate in estimating missing measurements, and that this accuracy is reflected in improved prediction of outcomes.
	
	The most familiar methods for estimating missing data follow one of three approaches, usually called {\em interpolation, imputation} and {\em matrix completion}.  Interpolation methods  such as \cite{interpolation,wavelet} exploit the correlation among measurements at different times {\em within each  stream} but ignore the correlation across streams. Imputation methods such as \cite{Rubin,EM,MICE,missforest} exploit the correlation among measurements at the same time {\em across different  streams} but ignore the correlation within streams.  Because medical measurements are frequently correlated both within streams {\em and} across streams (e.g., blood pressure at a given time is correlated both with blood pressure at other times and with heart rate), each of these approaches loses potentially important information. Matrix completion methods such as \cite{Mat-0,Mat-1,Mat-2,Mat-3} do exploit correlations  within and across streams, but assume that the data is static -- hence ignore the temporal component of the data -- or that the data is perfectly synchronized -- an assumption that is routinely violated in medical time series data.  Some of these methods also  make modeling assumptions about the nature of the data-generating process or of the pattern of missing data. Our approach is expressly designed to exploit both the correlation within streams and the correlation across streams and to take into account the temporal and non-synchronous character of the data; our approach makes no modeling assumptions about the data-generating process or the pattern of missing data. (We do assume -- as is standard in most of the literature -- that the data is {\em missing at random} \cite{Missing_Book}.  Dealing with data that is not missing at random \cite{ICML2017-Alaa} presents additional challenges and is left for future works.)  	
	
\begin{figure*}[t!]
		\center
		\includegraphics[width=0.9\textwidth]{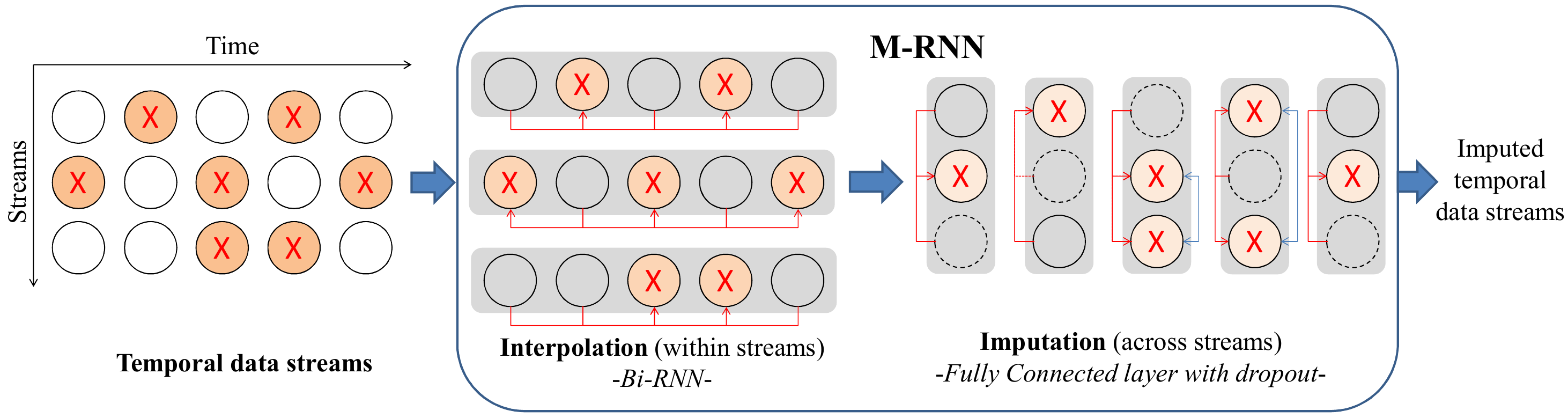}
		\caption{Block diagram of missing data estimation process: X = missing measurements; red lines = connections between observed values and missing values in each layer; blue lines = connections between interpolated values; dashed lines = dropout}
		\label{fig:High_Level}
	\end{figure*}

	Our method relies on a novel neural network architecture that we call a {\em Multi-directional Recurrent Neural Network (M-RNN)}.  Our M-RNN contains both an interpolation block and  an imputation block and it trains these blocks {\em simultaneously}, rather than separately (See Fig. \ref{fig:High_Level} and \ref{fig:Block_Time}). Like a bi-directional RNN (Bi-RNN) \cite{BiRNN}, an M-RNN operates forward and backward {\em within} each data stream -- in the {\em intra-stream directions}.  An M-RNN also operates {\em across} different data streams -- in the {\em inter-stream directions}. Unlike a Bi-RNN, the timing of inputs into the hidden layers of our M-RNN is lagged in the forward direction and advanced in the backward direction. As illustrated in Fig. \ref{fig:Block_Time}, our M-RNN architecture exploits the 3-dimensional nature of the dataset. 
	
	An important aspect of medical data is that there is often enormous uncertainty in the measured data.  As is well-known, although single imputation (SI) methods may yield the most plausible/most likely estimate for each missing data point \cite{singleimpute}, they do not capture the uncertainty in the imputed data \cite{Rubin}. Multiple imputation (MI) methods capture this uncertainty by sampling imputed values several times in order to form multiple complete imputed datasets, analyzing each imputed dataset separately and combining the results  via Rubin's rule  \cite{Rubin,Multipleimpute1,reason_of_imputation1}. Capturing the uncertainty in the dataset is especially important in the medical setting, in which diagnostic, prognostic and treatment decisions must be made on the basis of the imputed values \cite{Medimpute2, Medimpute3}.  In our setting, we use dropout \cite{Dropout} to produce multiple imputations; see section \ref{sect:MI}.
	
	To demonstrate the power of our method, we apply it to five different public real-world medical datasets: the MIMIC-III \cite{MIMIC} dataset, the clinical deterioration dataset used in \cite{TBME},  the UNOS  dataset for heart transplantation,  the UNOS dataset for lung transplantation (both available at  \underline{https://www.unos.org/data/}), and the UK Biobank dataset \cite{Biobank}.  We show that our method yields large and statistically significant improvements in estimation accuracy over previous methods, including  interpolation methods such as \cite{interpolation,wavelet}, imputation methods such as \cite{Rubin,EM,MICE,missforest}, RNN-based imputation methods such as \cite{Recent_RNN1,Recent_RNN2,Recent_RNN3} and matrix completion methods such as \cite{Mat-0}. For the MIMIC-III and clinical deterioration datasets the patient measurements were made frequently (hourly basis), and our method provides Root Mean Square (RMSE) improvement of more than 50\% over all 11 benchmarks.  For the UNOS heart and lung transplantation datasets and the Biobank dataset, the patient measurements were made much less frequently (yearly basis), but our method still provides RMSE improvement of more than 40\% in most cases, and significant improvements in the other cases.  We also show that this improvement in estimation  yields (smaller) improvements in the predictions of outcomes (patients' future states).  A number of experiments based on these same datasets show that the extent to which our method improves on outcomes depends on the method used for prediction, on the way in which our model is optimized in training, on the amount of data available (both in terms of the number of patients for whom we have data and on the amount of data available for each patient), and on the nature and extent of missing data.  These results illustrate the important point that, as mentioned earlier, there are {\em many} reasons for imputing missing data \cite{reason_of_imputation1,reason_of_imputation2} -- for the estimation of parameters (e.g. means or regression coefficients), for determination of confidence intervals and significance, as well as for prediction -- and that no single method for imputing data can be expected to be  superior {\em on all datasets} or {\em for all reasons }.
	
	As \cite{congeniality_meng} has emphasized, an extremely desirable aspect of any imputation method is that it be {\em congenial}; i.e. that it should produce imputed values in a manner that preserves the original relationships between features and labels.  As we demonstrate using the complete Biobank dataset, our method is also more congenial than the best competing benchmarks; see section \ref{sect:congenial}.

	\section{Related Work}	

    As we have noted, there are three  standard and very widely-used methods for dealing with missing data: interpolation, imputation and matrix completion. Interpolation methods \cite{interpolation,wavelet} attempt to reconstruct missing data by capturing the temporal relationship {\em within} each data stream but not the relationships {\em across} streams. Imputation methods \cite{Rubin,EM,MICE,missforest} attempt to reconstruct missing data by capturing the synchronous relationships {\em across} data streams but not the temporal relationships {\em within} streams. Matrix completion methods \cite{Mat-0,Mat-1,Mat-2,Mat-3} treat the data as static -- ignoring the temporal aspect -- or perfectly synchronized and assume a specific model of the data-generating process and/or the pattern of missing data.

	There is also a substantial  literature that uses Recurrent Neural Networks (RNNs) for prediction on the basis of time series with missing data.  For example, \cite{Early_RNN1} first replaces all the missing values with a mean value, then uses the feedback loop from the hidden states to update the imputed values and finally uses the reconstructed data streams as inputs to a standard RNN for prediction.  \cite{Early_RNN2} uses  the Expectation-Maximization (EM) algorithm to impute the missing values and again uses the reconstructed data streams as inputs to a standard RNN for prediction.   \cite{Early_RNN3} uses a linear model to estimate missing values from the latest measurement and the hidden state within each stream followed by a standard RNN for prediction.  In the first two of these papers, missing values are imputed by using only the synchronous relationships across data streams but not the  temporal relationships within streams; in the  third paper, missing values are interpolated by using only the temporal relationships within each stream but not on the relationships across streams.   
	
	A more recent literature extends these methods to deal with both missing data and irregularly sampled data \cite{Recent_RNN1,Recent_RNN2,Recent_RNN3,Recent_RNN5}. All of these papers use the sampling times to capture the informative missingness and time interval information to deal with irregular sampling, using the measurements, sampling information and time intervals as the inputs of an RNN. However, they differ in the replacements they use for missing values. \cite{Recent_RNN1,Recent_RNN2,Recent_RNN5} replace the missing values with 0, mean values or latest measurements -- all of which are independent of either the intra-stream or inter-stream relationships or both. \cite{Recent_RNN3} imputes the missing values using only the most recent measurements, the mean value of each stream, and the time interval.

	\section{Problem Formulation}\label{sect:model}
	Our formulation and method are applicable to a wide variety of settings with missing data.  However, for ease of exposition -- and to facilitate the discussion of our application to medical datasets -- it is convenient to adopt medical terminology throughout.
	
	We consider a dataset consisting of $N$  patients. For each patient, we have a multivariate time series data stream of length $T$ (the length $T$ and the other components of the dataset may depend on the patient $n$ but for the moment we suppress the dependence on $n$) that consists of time stamps $\mathcal{S}$, measurements $\mathcal{X}$, and labels $\mathcal{Y}$, sampled from an (unknown) underlying distribution $\mathcal{F}$:  $(\mathcal{S},\mathcal{X},\mathcal{Y}) \sim \mathcal{F}$. 
	
	For each $t$ the {\em time stamp} $s_t \in \mathbb{R}$ represents the {\em actual time} at which the measurements $x_t$ were taken. For convenience we normalize so that $s_1 = 0$ (so that we are measuring actual times for each patient beginning from the first observation for that patient); we assume actual times are strictly increasing: $s_{t+1} > s_t$ where $0 \leq t < T$. Note that the measurements may not be sampled regularly, so that the interval  $s_{t+1} - s_t$ between successive measurements need not be constant. 
	
	There are $D$ streams of measurements.  We view each measurement as a real number, but it will typically be the case that not every stream is actually observed/measured at $s_t$.  Hence we adopt notation in which the set of possible measurements at the $t$-th time stamp $s_t$ is $\mathbb{R}_* = \mathbb{R} \cup \{*\}$.  We interpret $x_t^d = *$  to mean that the stream $d$ was not measured at  $s_t$; otherwise $x_t^d \in \mathbb{R}$ is the actual measurement of stream $d$ at  $s_t$. For convenience, we scale all measurements to lie in the interval $[0,1]$.
	
	It is convenient to introduce some additional notation.  For each $t$, define the index $m^t_d$ to equal $0$ if $x_t^d = *$ (i.e. the stream $d$ was not measured at $s_t$) and to equal $1$ if $x_t^d \in [0,1]$ (the stream $d$ was measured at $s_t$).  We define  
	$\delta_t^d$ to be the actual amount of time that has elapsed from $s_t$ since the stream $d$ was measured previously; $\delta_t^d$ can be  defined by setting $\delta_1^d = 0$ and then proceeding recursively as follows: 
	\begin{equation*}
	\delta_t^d=\begin{cases}
	s_t - s_{t-1} + \delta_{t-1}^d & \text{if $t>1, m_{t-1}^d = 0$}.\\
	s_t - s_{t-1} & \text{if $t>1, m_{t-1}^d = 1$}\\
	\end{cases}
	\end{equation*}
	Write $\boldsymbol{\delta}_t$ for the vector of elapsed times at time stamp $t$ and $\Delta = \{ \boldsymbol{\delta}_1,\boldsymbol{\delta}_2,...,\boldsymbol{\delta}_T  \}$.
	
	The label $y_t $ represents the outcome realized at time stamp $t$ (actual time $s_t$) such as discharge, clinical deterioration, death. $\mathcal{Y}$ is the vector of outcomes for this patient. Again, we scale so the labels (and eventually predictions) lie in the interval $[0,1]$.  Frequently the outcome is binary in which case $y_t = 0$ or $y_t = 1$.
	
	The information available for  a particular patient $n$ is therefore a triple consisting of a sequence of time stamps, an array of measurements at each time stamp (with the above convention about missing measurements), and an array of labels at each time stamp. It is convenient to use functional notation to identify information about a particular patient, so $x_t^d(n)$ is the measurement of stream $d$ at time stamp $t$ for patient $n$, etc. The entire dataset consists of all the triples for all the patients 
	$\mathcal{D} = \{(\mathcal{S}(n), \mathcal{X}(n), \mathcal{Y}(n) \}_{n=1}^N$.
	
	Our   objective is to find a function $\mathbf{f}$ that provides the best estimate of missing values; i.e. the estimate that minimizes the estimation loss.  As is usually done, we measure loss as the squared error, so if $x_t^d$ is an (unobserved) actual measurement (sampled from 	$\mathcal{F}$) and $\hat{x}_t^d = f^d_t(\mathcal{S}, \mathcal{X})$ is the estimate formed on the basis of observed data, then the mean square loss for this particular measurement is ${\mathcal L}(\hat{x}_t^d, {x}_t^d) = (\hat{x}_t^d - x_t^d)^2$.  Hence the formal optimization problem is  to find a function $\mathbf{f}$ to solve:
	\begin{align}\label{eq:main_opt}
	\min_{\mathbf{f}} \, & \mathbb{E}_{\mathcal{F}} \Big[ \sum_{t=1}^T \sum_{d=1}^D (1-m_t^d) {\mathcal L}( \hat{x}_t^d, {x}_t^d ) \Big] \nonumber \\
	=\min_{\mathbf{f}}& \, \mathbb{E}_{\mathcal{F}} \Big[ \sum_{t=1}^T \sum_{d=1}^D (1-m_t^d) (f_t^d(\mathcal{S}, \mathcal{X}, \mathcal{Y}) - x_t^d)^2\Big].
	\end{align}
    Note that the function $f$ we seek depends on the particular  $d$ and $t$, and on the entire array of time stamps and measurements -- but not on labels (which may not be observed). Also note that the formal  problem asks to find an $\mathbf{f}$ that minimizes the loss with respect to the {\em true distribution}.  Of course we do not observe the true distribution and cannot compute the true loss, so we will minimize the empirical loss.

	\section{Multi-directional Recurrent Neural Networks (M-RNN)}
	Suppose that stream $d$ was not measured at time stamp $t$, so that $x_t^d = *$.  We would like to form an estimate 
	$\hat{x}_t^d$ of what the actual measurement would have been. As we have noted, familiar interpolation methods use only the measurements $x_{t'}^d$ of the fixed data stream $d$ for other time stamps $t' \neq t$ (perhaps both before and after $t$) -- but ignores the information contained in other data streams $d' \neq d$;  familiar imputation methods use only the measurements $x_t^{d'}$ at the fixed time $t$ for other data streams $d' \neq d$ -- but ignores the information contained at other times $t' \not=t$.  Because information is often correlated both {\em within and across} data streams, each of these familiar approaches throws away potentially useful information. Our approach forms an estimate 
	$\hat{x}_t^d$  using  measurements both {\em within the given data stream and  across other data streams}. In principle, we could try to form the estimate $\hat{x}_t^d$ by using {\em all} the information in $\mathcal{D}$. However, this would be impractical because it would require learning a  number of parameters that is on the order of the square of the number of data streams, and also because it would create a serious danger of over-fitting. Instead, we propose an efficient hierarchical learning framework using a novel RNN architecture that effectively allows us to capture the correlations both within streams and across streams.  Our approach limits the number of parameters to be learned to be of the linear order of the number data streams and avoids over-fitting. See Fig. \ref{fig:High_Level}.
	
\begin{figure}[t!]
    	\center
		\includegraphics[width=0.8\textwidth]{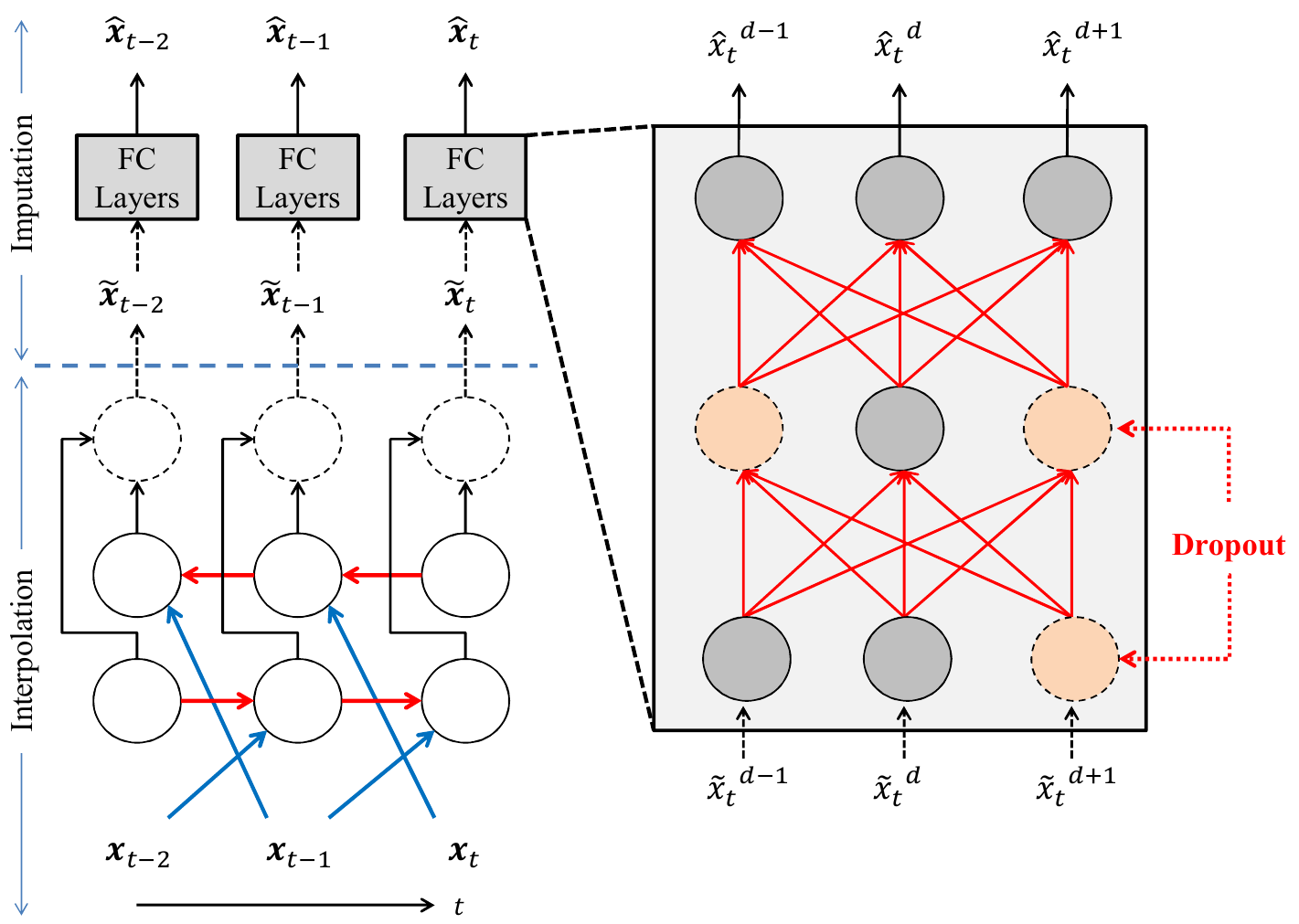}
		\caption{M-RNN Architecture. (Dropout is used for multiple imputations)}
		\label{fig:Block_Time}
	\end{figure}

	Our basic single-imputation M-RNN consists of 2 blocks: an Interpolation block  and an Imputation block; see Fig. \ref{fig:Block_Time}. (Our construction puts the Imputation block after the Interpolation block in order to use the outputs of the Interpolation block to improve the accuracy of the Imputation block; as we discuss later, it would not be useful to put the Interpolation block after the Imputation block.) To produce multiple imputations, we adjoin an additional dropout layer to the basic single-imputation M-RNN.  (We defer the details until Section \ref{sect:MI}.) 
	
	\subsection{Error/Loss } As formalized above  in Equation (\ref{eq:main_opt}), our overall objective is to minimize the error that would be made in estimating missing measurements. Evidently, we cannot estimate the error of a measurement that was not made and hence is truly missing in the dataset. Instead we fix a measurement ${x}_t^d$ that {\em was} made and is present in the dataset, form an estimate $\hat{x}_t^d$ for ${x}_t^d$ using only the dataset with ${x}_t^d$ removed (which we denote by $\mathcal{D} - {x}_t^d$), and then compute the error between the estimate $\hat{x}_t^d$ and the actual measurement ${x}_t^d$.  As above, we use the squared error $(\hat{x}_t^d - x_t^d)^2$ as the loss for this particular estimate; as  the total  loss/error for the entire dataset $\mathcal{D}$ we use the {\em mean squared error} (MSE):
	\begin{align*}
	{\mathcal L}(\{ \hat{x}_t^d, {x}_t^d \})  =\sum_{n=1}^N \left[ \frac{ \sum_{t = 1}^{T_n} \sum_{d = 1}^D m_t^d(n) \times (\hat{x}_t^d(n) - x_t^d(n))^2}{\sum_{t = 1}^{T_n} \sum_{d = 1}^D m_t^d(n) } \right]
	\end{align*}
    Note that this is the empirical error, which is actually achievable.
		
	\subsection{Interpolation Block } The Interpolation block constructs an interpolation function 
	$\boldsymbol{\Phi}$ that operates {\em within} a given stream. To emphasize that the estimate $\tilde{x}^d_t$ for $x^d_t$ depends on the data with $x^d_t$ removed, we write $\tilde{x}^d_t = \boldsymbol{\Phi}(\mathcal{D} - {x}_t^d)$; but keep in mind that the construction uses  only  the data from stream $d$,  not the data from other streams.  We construct
	$\boldsymbol{\Phi}$ using a bi-directional recurrent neural network (Bi-RNN).  However, unlike a conventional Bi-RNN \cite{BiRNN}, the timing of inputs into the hidden layer is lagged in the forward direction and advanced in the backward direction: at $t$,  inputs of forward hidden states come from $t-1$ and inputs of backward hidden states come from $t+1$. (This procedure ensures that the actual value $x^d_t$  is automatically not used in the estimation of $\tilde{x}^d_t$.) If we write $\textbf{z}_t^d = [x^d_t, m^d_t, \delta^d_t]$ then a more mathematical description is:
	\begin{align*}
	\tilde{x}^d_t &= g(U^d [\overrightarrow{\textbf{h}}_t^d ;\overleftarrow{\textbf{h}}_t^d] + \textbf{c}_o^d) = g(\overrightarrow{U}^d \overrightarrow{\textbf{h}}_t^d + \overleftarrow{U}^d \overleftarrow{\textbf{h}}_t^d + \textbf{c}_o^d) \\
	\overrightarrow{\textbf{h}}_t^d &= f(\overrightarrow{W}^d \overrightarrow{\textbf{h}}_{t-1}^d + \overrightarrow{V}^d \textbf{z}_{t-1}^d + \overrightarrow{\textbf{c}}^d) \\
	\overleftarrow{\textbf{h}}_t &= f(\overleftarrow{W}^d \overleftarrow{\textbf{h}}_{t+1}^d + \overleftarrow{V}^d \textbf{z}_{t+1}^d + \overleftarrow{\textbf{c}}^d)
	\end{align*}
	Here, $f,g$ are  activation functions.  (In principle, any  activation functions, such as ReLu, tanh, etc., could be used; here we use ReLu.)  The arrows indicate forward/backward direction.  As we have emphasized, in this  interpolation block, we are only using/capturing the temporal correlation {\em within each data stream}. In particular, the parameters for each data stream are learned separately, and the number of parameters that must be learned is linear in the number of streams $D$. (The weight matrices $W, U, V$ are block diagonal.) 
	
	\subsection{Imputation Block } The  Imputation blocks constructs an imputation function $\boldsymbol{\Psi}$ that operates {\em across} streams. To again emphasized that the estimate $\hat{x}^d_t$ for  ${x}_t^d$ depends on the data with ${x}_t^d$ removed, we write $\hat{x}^d_t = \boldsymbol{\Psi}(\mathcal{D} - {x}_t^d)$; again, keep in mind that now we are using only data at time stamp $s_t$, not data from other time stamps. We construct the function $\boldsymbol{\Psi}$ to be independent of $t$, so we use fully connected layers; see the Imputation component of Fig \ref{fig:Block_Time}. If we write $\textbf{z}_t = [\tilde{\textbf{x}}_t, \textbf{m}_t]$ then a more mathematical description is:	
	\begin{align*}
	\hat{\textbf{x}}_t &= \sigma(W \textbf{h}_t + \alpha) &
	\textbf{h}_t = \phi(U \textbf{x}_t + V \textbf{z}_t + \beta)
	\end{align*}
	where $\sigma, \phi$ are  activation functions.  The diagonal entries of  $U$ are zero because we do not use $x_t^d$ to estimate $\hat{x}_t^d$.   
		
	We learn the functions $\boldsymbol{\Phi}$ and $\boldsymbol{\Psi}$ {\em jointly} using the stacked networks of Bi-RNN and Fully Connected (FC) layers, using MSE as the objective function.
	\begin{align}\label{eq:emp_opt}
	\boldsymbol{\Phi}^*,\boldsymbol{\Psi}^* =\arg\min_{\boldsymbol{\Phi},\boldsymbol{\Psi}}   \mathcal{L}(\{ \boldsymbol{\Psi} \Big(\{ x_t^d,\boldsymbol{\Phi} \Big(\{ x_t^d,m_t^d,\delta_t^d  \}_{t = 1}^T\Big),m_t^d  \}_{d = 1}^D\Big), x_t^d \})
	\end{align}

	\subsection{Multiple Imputations} \label{sect:MI} 
	It is well-understood that to account for the uncertainty in estimating missing values, it is useful to produce multiple estimates and generate multiple imputed datasets. These multiple imputed datasets can each be analyzed using standard methods and the results can be combined using Rubin’s rule \cite{Rubin}. In our case, we generate multiple imputed datasets using the well-known Dropout \cite{Dropout} approach: we randomly selects neurons in the fully connected layers and delete those neurons and all their connections. (The dropout probability $p \in (0,1)$ is a hyper-parameter to be chosen; the neurons to be dropped are chosen according to the Bernoulli distribution with parameter $p$.) In the training stage, we conduct joint optimization (equation (\ref{eq:emp_opt})) using the dropout process. We then generate multiple outputs $\textbf{o}_t$ by sampling different dropout vectors $\textbf{R}$ from the Bernoulli distributions. This yields multiple imputations (MI). (To construct a single imputation (SI) we proceed in precisely the same way but set the dropout probability to $0$.  For comparisons, we normalize the final output by multiplying by $p$.)
	
	\subsection{Overall Structure } We refer to the entire structure above as a {\em Multi-directional Recurrent Neural Network (M-RNN)}.   We use the notations M-RNN (MI) and M-RNN (SI) to clarify whether we are producing multiple or single imputations.

	\section{Datasets}\label{sect:dataset}
	We use five datasets, the characteristics of which are summarized in Table \ref{tab:Dataset}: more detailed descriptions are below.

	\begin{table*}[t!]
		\caption{Summary of the datasets (Cont: Continuous, Cat: Categorical, Avg: Average)}
		\label{tab:Dataset}
		\centering
		\small
		\begin{tabular}{c|c|c|c|c|c}
			\toprule
			\textbf{Datasets} & \textbf{MIMIC-III} & \textbf{Deterioration} & \textbf{UNOS-Heart} & \textbf{UNOS-Lung} &\textbf{Biobank}\\
			\midrule
			\textbf{Number of Patients} & 23,160& 6,094&69,205& 32,986 & 3,902\\
			\midrule
			\textbf{Number of Dimensions (Cont, Cat)} &40 (31, 9) & 38 (16, 22) & 34 (10, 24) & 34 (10, 24) & 113 (67, 46)\\		
			\midrule
			{\textbf{Label} $(y = 1)$} &1,320 (5.7\%) &306 (5.3\%) & 4,844 (7.0\%) & 2,276 (6.9\%) & 195 (5.0\%)\\
			\midrule
			\textbf{Avg number of samples} & 24.3 & 34.3 & 6.2 & 4.0 & 3.0\\
			\midrule
			\textbf{Avg missing rate} & 75.0\% &61.4\%& 59.1\% & 58.5\% & 0.0\% \\
			\midrule
			\textbf{Avg Measurement freq.} & 1 hr / 12 hrs & 4 hrs / 24 hrs&1 year & 1 year & 2.3 years\\
			\midrule
			\textbf{Avg Correlation within streams}  & {0.4122} & {0.3436} & {0.1213}&{0.1157} & {0.2424}\\
			\midrule
			\textbf{Avg Correlation across streams}  & {0.3127} & {0.3454} &{0.0875}&{0.0897} & {0.0506}\\
			\bottomrule
		\end{tabular}
	\end{table*}

	\subsection{MIMIC-III} The dataset MIMIC-III \cite{MIMIC} contains data for patients monitored in intensive care units (ICUs) of various hospitals. Within the entire MIMIC-III dataset, we only used the data for the 23,160 patients whose measurements are recorded by Metavision (post 2008). We used 20 vital signs (e.g. heart rate, respiratory rate, blood pressures, etc.) and 20 lab tests (e.g. creatinine, chloride, etc.) whose missing rates are minimum. For these patients we have 40 physiological data streams in all. Vital signs were sampled roughly every hour; lab tests were sampled roughly every 12 hours. Each patient was followed until either death (1,320 patients (5.7 \%)) or discharge from ICU (21,840 patients (94.3 \%)). Note that because lab tests are sampled only 1/12 as often as vital signs, in effect 11/12 of lab test data is missing -- even if every lab test for every patient was actually conducted and recorded. For the purpose of prediction, we take the goal as predicting at each time $t$ whether the patient will die within the next 24 hours.  Hence we assign the label $y_t = 1$ of the patient actually died within the 24 hrs following the time $s_t$ and  $y_t = 0$ otherwise.
	
	\subsection{Deterioration} The dataset described in \cite{TBME} provides records for a cohort of 6,094 patients who were followed for potential clinical deterioration while hospitalized. Patients were monitored for 28 vital signs (heart rate, blood pressure, etc.) and 10 lab tests (creatinine, hemoglobin, etc.) so there are 38 physiological data streams in all. Vital signs were sampled roughly every 4 hours; lab tests were sampled roughly every 24 hours. Each patient was followed until either admission to ICU (306 patients (5.0 \%)) or discharge from hospital (5,788 patients (95.0\%)). Again, because lab tests are sampled only 1/6 as often as vital signs, in effect 5/6 of lab test data is missing -- even if every lab test for every patient was actually conducted and recorded. For the purpose of prediction, we take the goal as predicting at each time stamp $s_t$ whether the patient will be admitted to the ICU (experienced clinical deterioration) within the next 24 hours.  Hence we assign the label $y_t = 1$ if the patient was admitted to the ICU within the following 24 hours and $y_t = 0$ otherwise.

	\subsection{UNOS-Heart and UNOS-Lung} The UNOS (United Network for Organ Transplantation) dataset (available at  \underline{https://www.unos.org/data/}) provides yearly  follow-up information for the entire U.S. cohort of 69,205 patients who received heart transplants and 32,986 patients who received lung transplants during the period 1985-2015. We view patients in the dataset as described by a total of 34 clinical features. (In fact, a total of 232 features are recorded in the UNOS dataset, but many features are not recorded for most patients.  We therefore excluded the 198 features for which missing rates were higher than 80\%; this is in keeping with standard medical statistical practice.) For each patient, a number of yearly follow-ups are recorded; the smallest number of yearly follow-ups is 1, the largest is 26; the median number of follow-ups for heart transplantation is 6 and the median number of follow-ups for lung transplantation is 4.  (In the main text, we focus entirely on features of the patient, ignoring features of the donor.  We do this for two reasons: (i) the features of the patient change over time but the features of the donor do not (because the donor is dead); (ii) the relevant features of the donor appear to be largely captured in the time series measurements of the patient.  However, as we show in  the Appendix, taking features of the donor into account seems to make little difference for either imputation or prediction.) For the purpose of prediction, we take the goal in each case as predicting at each follow up time $s_t$ whether the patient will be dead one year later. Hence we assign the label $y_t = 1$ if the patient actually died within the following year and $y_t = 0$ otherwise.

	\subsection{Biobank} We used the UK Biobank dataset gathered from 21 assessment centers across England, Wales, and Scotland using standardized procedures from 2007 to 2014. (The UK Biobank protocol is available online.) UK Biobank recorded various patient information including baseline measurements, physical measurements, and evaluations of biological samples. For this paper, we excluded all the variables that were missing for more than 80\% of the participants and all the static measurements and only used the 113 longitudinal measurements.  Of the 4,096 total patients, we used only the data for the 3,902 patients who missed no admissions to assessment centers (and hence were assessed the maximum number of times, which was three) and for whom there are no missing measurements of these 113 variables; thus we have a complete dataset.  For the purpose of prediction we take the goal as the correct prediction of diabetes, so we assign the label $y_t = 1$ if diabetes is diagnosed at $t$ and $y_t = 0$ otherwise.

    \section{Results and Discussions}
    \subsection{Imputation Accuracy on the Given Datasets}  
    We begin by comparing the performance of our method (using both multiple imputations and single imputation) on the  given datasets against 11 benchmarks with respect to the accuracy of imputing missing values. The benchmarks against which we compare are: the algorithms proposed in \cite{Recent_RNN1,Recent_RNN2,Recent_RNN3}; Spline and Cubic Interpolation \cite{interpolation} ; MICE \cite{MICE}; MissForest \cite{missforest}, EM \cite{EM}; the matrix completion algorithm of \cite{Mat-0};the Auto-Encoder algorithm proposed in \cite{autoencoder}; and the Markov chain Monte Carlo (MCMC) method \cite{mcmc}. (The details of the implementations for the various benchmarks are presented in the Appendix.)  As is common, we use root mean square error (RMSE) as the measure of performance. In each experiment, we use 5 cross-validation. Table \ref{tab:Imputation} shows the mean RSME for our method and benchmarks, and the percentage improvement of RMSE for M-RNN (MI) over the benchmarks. (Note that we are unable to provide results for the EM algorithm on the UNOS-Heart and UNOS-Lung datasets because -- at least for the implementation we use -- the EM algorithm requires at least one patient for whom data is complete, and the UNOS-Heart and UNOS-Lung datasets do not contain any such patient.)

     \begin{table*}[t!]
	\caption{Performance comparison for missing data estimation }
	\label{tab:Imputation}
	\centering
	\small
	\begin{tabular}{c|c|c|c|c|c|c}
		\toprule
		\multirow{2}{*}{\textbf{Category}}&\multirow{2}{*}{\textbf{Algorithm}}&\multicolumn{5}{c}{\textbf{Mean RMSE (\% Gain of M-RNN (Multiple Imputations))}}  \\
		\cmidrule{3-7}
		& & MIMIC-III & Deterioration & UNOS-Heart & UNOS-Lung & Biobank \\
		\midrule
		\multirow{2}{*}{\textbf{M-RNN}}&\textbf{M-RNN (MI)} &\textbf{0.0141}  \textbf{(-)} &  \textbf{0.0105}  \textbf{(-)} &\textbf{0.0479}  \textbf{(-)}&\textbf{0.0606}  \textbf{(-)} &\textbf{0.0637}  \textbf{(-)}\\
		& \textbf{M-RNN (SI)} &\textbf{0.0144}  \textbf{(-)}&  \textbf{0.0108}  \textbf{(-)}  &\textbf{0.0477}  \textbf{(-)}&\textbf{0.0609}  \textbf{(-)}&\textbf{0.0629}  \textbf{(-)} \\
		\midrule
		\multirow{3}{*}{\textbf{RNN-based}}&\cite{Recent_RNN1} &0.0337  (58.2\%)& 0.0258  (59.3\%) & 0.1352 (64.6\%)&   0.1343 (54.9\%) &   0.0812 (21.6\%) \\
		&\cite{Recent_RNN2} &0.0295   (52.2\%)& 0.0241   (56.4\%)& 0.1179 (59.4\%) & 0.1264 (52.1\%)&   0.0801 (20.5\%)\\
		&\cite{Recent_RNN3} &0.0292   (51.7\%)& 0.0233   (54.9\%)& 0.1057 (54.7\%) & 0.1172 (48.3\%)&   0.0778 (18.1\%) \\
		\midrule
		\multirow{2}{*}{\textbf{Interpolation}} &Spline &0.0735  (80.8\%)  &  0.0215  (51.2\%)& 0.1102  (56.5\%) & 0.1199 (49.5\%)&   0.0845 (24.6\%)\\
		&Cubic   &0.0279  (49.5\%)& 0.0223  (52.9\%)& 0.1072 (55.3\%) & 0.1177 (48.5\%)& 0.0887 (28.2\%)\\
		\midrule
		\multirow{3}{*}{\textbf{Imputation}}&MICE &0.0611 (76.9\%)&  0.0319 (67.1\%) &0.1147  (58.2\%)& 0.1151  (47.4\%)& 0.0915 (30.4\%)\\
		& MissForest & 0.0293 (51.9\%)& 0.0264 (60.2\%)  &0.0489  (2.0\%)& 0.0652  (7.1\%) & 0.0892 (28.6\%) \\
		& EM  &0.0467  (69.8\%)&  0.0355  (70.4\%)&- & -& 0.0978 (34.9\%) \\ \midrule
		\multirow{3}{*}{\textbf{Others}} &Matrix Completion  &0.0311  (54.7\%) &  0.0264  (60.2\%)&0.0974  (50.8\%)& 0.0942 (35.7\%) & 0.0886 (28.1\%) \\
		&Auto-encoder & 0.0412 (66.0\%) & 0.0309 (65.0\%)  & 0.0589 (18.7\%) & 0.0712 (14.9\%)& 0.0805 (20.9\%) \\	
		&MCMC  & 0.0437 (67.7\%) & 0.0364 (71.2\%)  & 0.1091 (56.1\%) & 0.1124 (46.1\%)& 0.0936 (31.9\%) \\			
		\bottomrule
	\end{tabular}
\end{table*}

As can be seen in Table \ref{tab:Imputation}, M-RNN achieves better performance (smaller RMSE) than all of the benchmarks on all of the datasets (for all comparisons are possible).  With a single exception (the comparison with MissForest on the UNOS-Lung dataset) the performance improvements are statistically significant at the 95\% level (i.e., $p < 0.05$), and many of the improvements are very large.  For instance, for the Deterioration dataset, M-RNN using multiple imputations achieves RMSE of 0.0105 (95\% CI: 0.0071-0.0138), while the {\em best} benchmark (Spline interpolation)  achieves RMSE of 0.0215 (95\% CI: 0.0178-0.0255); this represents an improvement of 51.2\%.

The performance comparisons across datasets are revealing, if not necessarily surprising. The interpolation benchmarks (such as Spline, Cubic and RNN-based methods) work best on datasets, such as MIMIC-III and Deterioration, for which measurements were more frequent (and  more highly correlated within each stream (see Table \ref{tab:Dataset})); the imputation benchmarks work best on datasets, such as UNOS-Heart and UNOS-Lung, for which measurements were less frequent but for which there were many streams of data (many dimensions). The improvement of our method over all benchmarks is larger for the MIMIC-III and Deterioration datasets because those datasets have many streams of frequently sampled data, so that our method gains a great deal from exploiting both the correlations within each data stream and the correlations across data streams. Conversely, the improvement of our method is smaller for the UNOS-Heart and UNOS-Lung datasets, because streams in those datasets are infrequently sampled to that there is less to be gained by exploiting the correlations within data streams.  (The performances of the benchmarks for the Biobank dataset are mixed, and don't quite fit this same pattern, perhaps because Biobank is a small dataset (less than 4,000 patients with complete temporal data streams).)

\subsubsection{Multiple Imputations vs. Single Imputation} 
\begin{figure}[t!]
	\centering
	\includegraphics[width=0.8\textwidth]{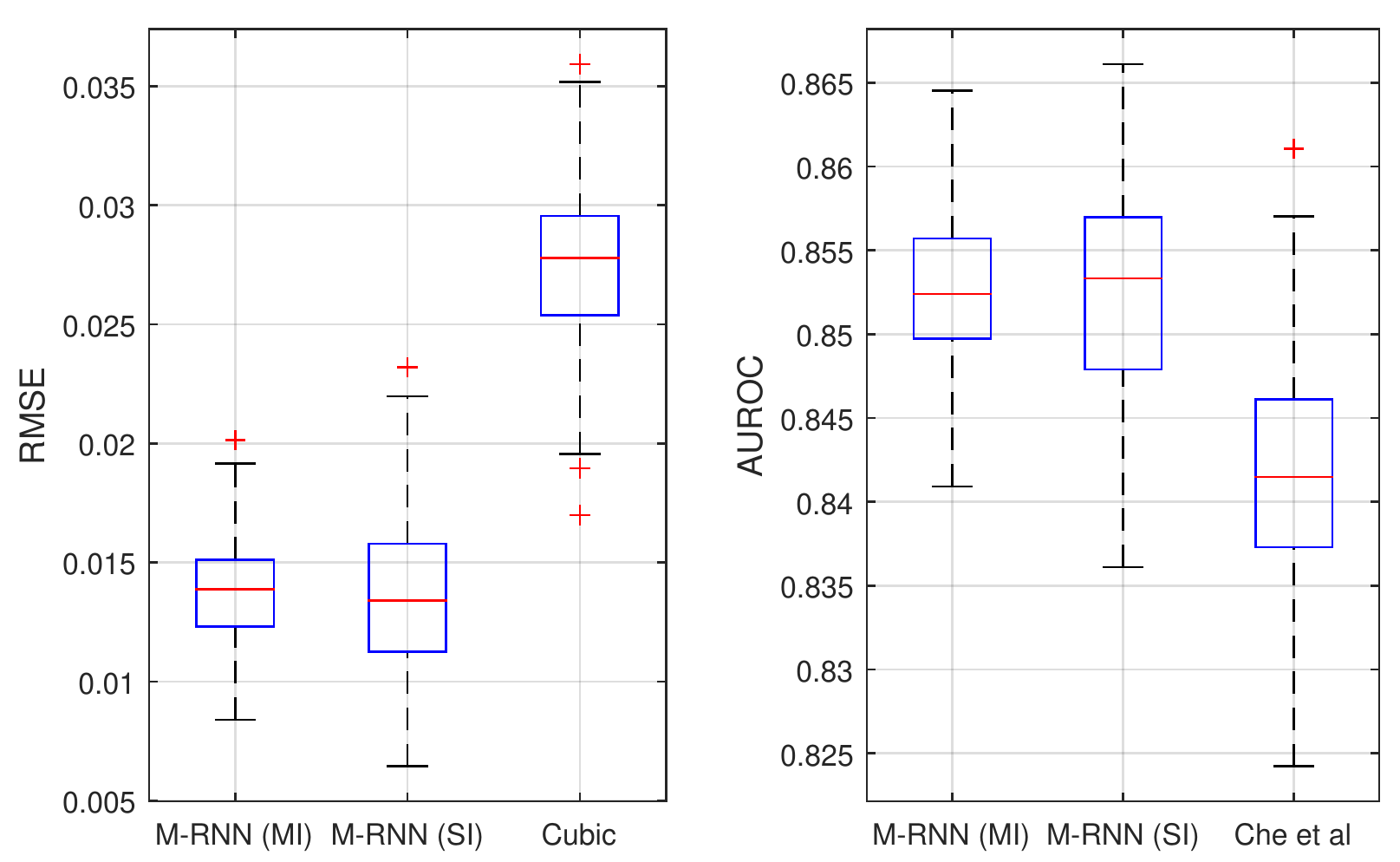}
	\caption{Box-plot comparisons between M-RNN (MI), M-RNN (SI) and the best benchmark. (a) RMSE comparison using MIMIC-III dataset, (b) AUROC comparison using MIMIC-III dataset}
	\label{fig:boxplot}
\end{figure}

As we have noted, the purpose of conducting multiple imputations is to reduce uncertainty/shrink confidence intervals (rather than to improve average performance).  As is illustrated in the box-plot in Fig. \ref{fig:boxplot}(a) which shows the comparison of M-RNN with multiple imputations and M-RNN with a single imputation against the best benchmark (Cubic interpolation) on the MIMIC-III dataset, our multiple imputations do achieve this purpose.  (For discussion of Fig. \ref{fig:boxplot}(b), which illustrates the corresponding reduction in uncertainty for prediction, see below.)

\subsubsection{Combining models of interpolation and imputation}
As we have already discussed, standard interpolation algorithms cannot capture the patterns across streams and standard imputation algorithms  cannot  capture the patterns within the streams.  However, it is possible to  combine a standard interpolation algorithm and standard imputation algorithm in an attempt to capture {\em both patterns}, and it might be thought that such a combination would be a fairer benchmark against which to compare our method.   To put this idea to the test, we create a family of ``joint algorithms'' by first using an interpolation algorithm to interpolate the missing values, and then using the interpolated values as the initial points of an imputation algorithm to provide final imputed values. For this exercise, we use two standard interpolation methods (Cubic and Spline), and two standard imputation methods (MICE and MissForest) so that we have 4 interpolation-imputation combination models: Cubic + MICE, Cubic + MissForest, Spline + MICE, and Spline + MissForest. 

\begin{table*}[t!]
	\caption{Performance comparison for joint interpolation/imputation algorithms}
	\label{tab:Sup_Imputation}
	\centering
	\small
	\begin{tabular}{c|c|c|c|c|c}
		\toprule
		\multirow{2}{*}{\textbf{Algorithm}} &\multicolumn{5}{c}{\textbf{Mean RMSE (\% Gain from Imputation Algorithm)}}  \\
		\cmidrule{2-6}
		 & MIMIC-III & Deterioration & UNOS-Heart & UNOS-Lung & Biobank \\
		\midrule
		Spline + MICE  & 0.0602 (1.5\%)  &  0.0320 (-0.3\%) & 0.1141 (0.5\%)  & 0.1133 (1.7\%)  & 0.0895 (2.2\%)\\
		Spline + MissForest   &0.0291 (0.7\%)  & 0.0259 (1.9\%)  & 0.0491 (-0.4\%)  & 0.0641 (1.4\%) & 0.0879 (4.1\%)\\
		Cubic + MICE & 0.0605 (1.0\%)&  0.0315 (1.3\%)  &0.1137 (0.9\%)  & 0.1138 (1.1\%) & 0.0901 (1.6\%) \\
		Cubic + MissForest & 0.0289 (1.4\%) & 0.0261 (1.1\%)   &0.0493 (-0.8\%)  & 0.0643 (1.4\%) & 0.0887 (3.2\%)  \\	\bottomrule
	\end{tabular}
\end{table*}

As Table \ref{tab:Sup_Imputation} shows, however, the  performances of these interpolation-imputation combination models are very similar to those of the performance of the simple imputation model that is used.  Indeed, the largest RMSE performance improvement is only $0.0018$.  The  reason for this is that imputation methods use algorithms that operate iteratively until they converge, so that their performance is rather robust to the initialization. Hence, although the interpolation part of the joint models captures some of the inter-stream information, the iterative imputation part ignores most of what is captured. 

\subsection{Source of Gains}
As illustrated in Fig. \ref{fig:Block_Time}, our M-RNN consists of an Interpolation block and an Imputation block. To understand where the gains of our approach come from, we compare the performance of that is achieved when we use only the Interpolation block or only the Imputation block; the results are shown in Table \ref{tab:sourceofgain}.

\begin{table}[t!]
	\caption{Source of Gain of M-RNN. (Performance degradation from original M-RNN) }
	\label{tab:sourceofgain}
	\centering
	\small
	\begin{tabular}{c|c|c|c}
		\toprule
		\multirow{2}{*}{\textbf{Datasets}} & \multicolumn{3}{c}{\textbf{M-RNN (Mean RMSE; \% Gain)}}\\
		\cmidrule{2-4}
		& \textbf{Only Interp} & \textbf{Only Impute} & \textbf{Interp + Impute} \\
		\midrule
		\multirow{2}{*}{\textbf{MIMIC-III}} & 0.0191 & 0.0312 & 0.0141 \\
		& (26.2 \%) & (54.8 \%) & (-) \\
		\midrule
		\multirow{2}{*}{\textbf{Deterioration}} & 0.0133 & 0.0295 & 0.0105 \\
		& (21.1 \%) & (64.4 \%) & (-) \\
		\midrule
		\multirow{2}{*}{\textbf{UNOS-Heart}} & 0.0897 & 0.0531 & 0.0479 \\
		& (46.6 \%) & (9.8 \%) & (-) \\
		\midrule
		\multirow{2}{*}{\textbf{UNOS-Lung}} & 0.0998 & 0.0734 & 0.0606 \\
		& (39.3 \%) & (17.4 \%) & (-) \\
		\midrule
		\multirow{2}{*}{\textbf{Biobank}} & 0.0794 & 0.0778 & 0.0637 \\
		& (19.8 \%) & (18.1 \%) & (-) \\
		\bottomrule
	\end{tabular}
\end{table}

The Interpolation block is intended to exploit the correlations within each data stream and the Imputation block is intended to exploit the correlations across streams, so it is to be expected that the largest gains of our M-RNN method should come from the Interpolation block for the datasets (MIMIC-III and Deterioration) which are frequently samples and have large temporal correlations, and should come from the Imputation block for the datasets (UNOS-Heart and UNOS-Lung) which are infrequently sampled but have many data streams.  As shown in Table \ref{tab:sourceofgain}, these intuitions are indeed supported by the experiments.  

\subsection{Additional Experiments}
\begin{figure*}[t!]
	\centering
	\includegraphics[width=0.7\textwidth]{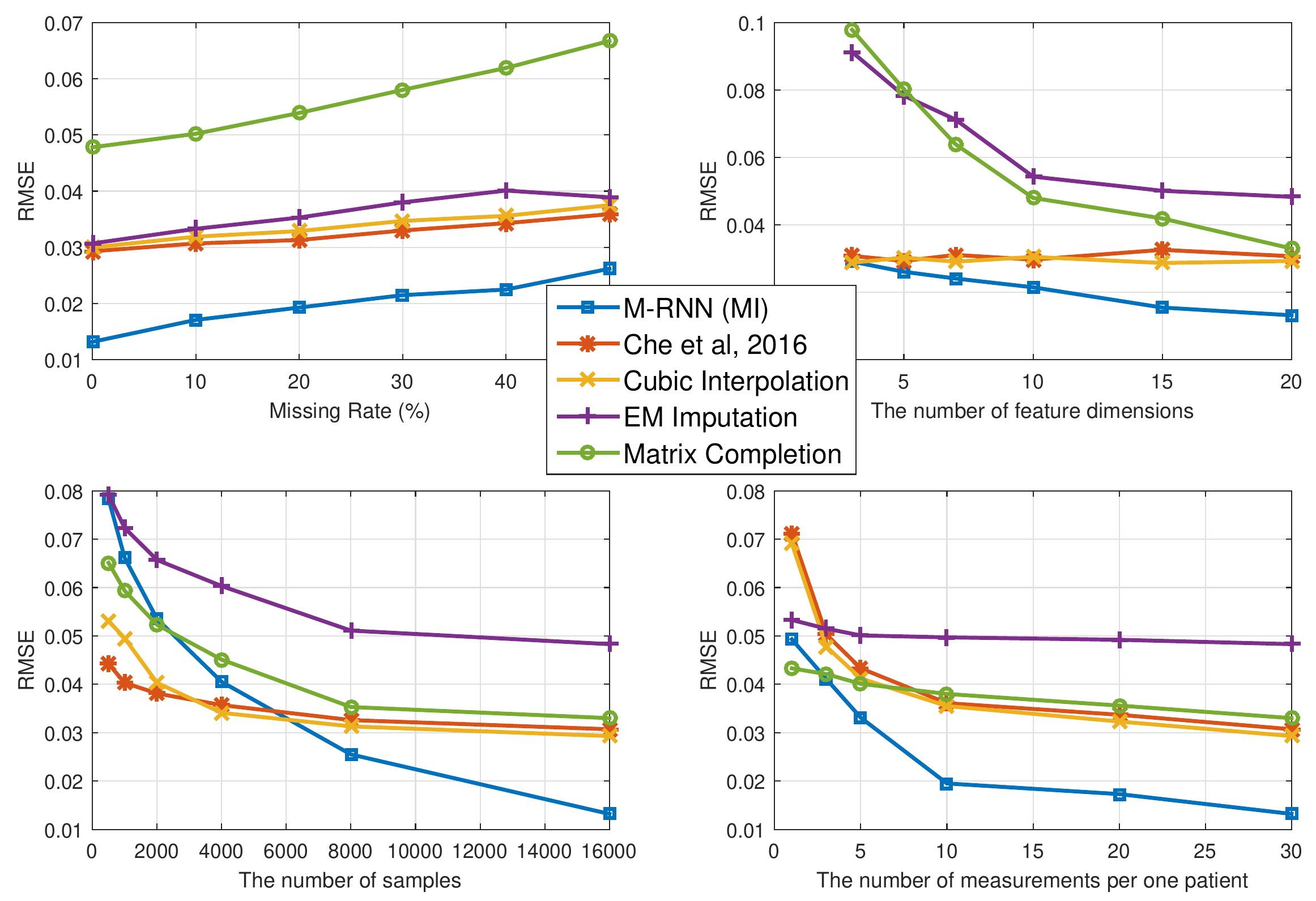}
	\caption{Imputation accuracy for the MIMIC-III dataset with various settings (a) Additional data missing at random (top-left), (b) Feature dimensions chosen at random (top-right), (c) Samples chosen at random (bottom-left), (d) Measurements chosen at random  (bottom-right)}
	\label{fig:various}
\end{figure*}

The experiments we have described above demonstrate that our method significantly outperforms a wide variety of benchmarks for the imputation of missing data on five somewhat representative datasets.  However it is natural to ask how our method would compare in other circumstances.  To get some understanding of this, we conducted four sets of experiments based on the MIMIC-III dataset: increasing the amount of missing data, reducing the number of data streams, reducing the number of samples, and reducing the number of measurements per patient.  Within each set of experiments, we conducted 10 trials for each value of the parameter being studied (e.g. amount of missing data), and we report the average over these 10 trials. The results are described below and in Fig. \ref{fig:various}. Although the results of these experiments are extremely suggestive, we caution the reader that these are only a specific set of experiments and that one should be careful about drawing general conclusions.  

\subsubsection{Amount of Missing Data}  To evaluate the performance of M-RNN in comparison to benchmarks in settings with more missing data, we constructed sub-samples of the MIMIC-III dataset by randomly removing 10\%, 20\%, 30\%, 40\%, 50\% of the actual data and carrying out the same estimation exercise as above on the smaller datasets that remain. (Recall that in the original MIMIC-III dataset, 75\% of the data is already missing; hence removing 50\% of the data present leads to an artificial dataset in which 87.5\% of the data is missing.)  The graph in Fig. \ref{fig:various}(a) shows the performance of M-RNN against the {\em best} benchmarks of each type for these smaller datasets.  As can be seen, M-RNN continues to substantially outperform the benchmarks.  Note that as the amount of missing data increases the improvement of M-RNN over the imputation benchmark(s) increases, but the improvement over the interpolation benchmarks decreases. 

\subsubsection{Number of Data Streams} As we have noted, typical medical datasets contain many data streams (many feature dimensions).  To evaluate the performance of M-RNN in comparison to benchmarks in settings with fewer data streams, we conducted experiments in which we reduced the number of data streams (feature dimensions) of MIMIC-III.  In the original MIMIC-III dataset the number of data streams is $D = 40$; we conducted experiments with  $D = 3,5,7,10,15,20$ data streams.  (In each case, we conducted 10 trials in which we selected data streams at random; we report the average of these 10 trials.)   As expected, the performance of M-RNN degrades when there are fewer data streams, but as Figure \ref{fig:various}(b) shows, M-RNN still outperforms the benchmarks.  (Note that interpolation methods are insensitive to the number of data streams because they operate only {\em within each data stream separately}.)

\subsubsection{Number of Samples} The original MIMIC-III dataset has $N = 23,160$ samples (patients).  To understand the performance of M-RNN in comparison to benchmarks in settings with fewer samples, we conducted experiments in which we used only subsets of all patients (samples) of sizes $N = 500, 1000, 2000, 4000, 8000,16000$.   Because M-RNN has to learn many parameters, it should come as no surprise that, as Fig. \ref{fig:various}(c) shows, the performance of M-RNN degrades badly -- and indeed is worse than that of (some) other benchmarks -- when the number of samples is too small, but M-RNN outperforms all the benchmarks as soon as the  number of training samples exceeds $N = 7,000$.  (However, one should not necessarily take the figure $N = 7,000$ as representing a cut-off below which M-RNN should not be applied, because  M-RNN outperforms the benchmarks on the Deterioration and Biobank datasets, which contain only 6,094 samples  and 3,902 samples, respectively.)

\subsubsection{Number of Measurements per Patient}  We have already noted that, in our datasets, MIMIC-III and Deterioration have many (relatively frequent) measurements per patient, while the other datasets have only a few (and infrequent) measurements per patient and that this leads to differences in performance of M-RNN. To further explore this effect, we created subsets of the MIMIC-III dataset with $T = 1, 3, 5, 10, 20, 30$ measurements per patient. As might be expected, and as Fig. \ref{fig:various}(d) shows, having fewer measurements per patient degrades the performance of  interpolation based algorithms but has little effect on pure imputation based methods; the performance of M-RNN is also degraded, but to a much lesser extent.

\subsection{Prediction Accuracy }  
As we have noted, there are many reasons for imputing missing data; one such is to improve predictive performance.  We therefore compare our method against the same 11 benchmarks with respect to the accuracy of predicting labels.  (See the description of the datasets in section \ref{sect:dataset} for labeling in each case.)  For this purpose, we use Area Under the Receiver Operating Characteristic Curve (AUROC) as the measure of performance. To be fair to all methods of imputing missing values, we use the same predictive model (a simple 1-layer RNN) in all cases.

\subsubsection{Prediction accuracy on the original datasets}
In this subsection, we evaluate the effects of the imputations on the prediction of labels (outcomes), which in the cases at hand correspond to prognoses.

\begin{table*}[t!]
	\caption{Performance comparison for patient state prediction with a 1-layer RNN}
	\label{tab:Prediction}
	\centering
	\small
	\begin{tabular}{c|c|c|c|c|c|c}
		\toprule
		\multirow{2}{*}{\textbf{Category}}&\multirow{2}{*}{\textbf{Algorithm}}&\multicolumn{5}{c}{\textbf{AUROC (\% Gain of M-RNN (Multiple Imputations)}}  \\
		\cmidrule{3-7}
		& & MIMIC-III & Deterioration & UNOS-Heart & UNOS-Lung & Biobank \\
		\midrule
		\multirow{2}{*}{\textbf{M-RNN}}&\textbf{M-RNN (MI)} &  \textbf{0.8531}  \textbf{(-)}  &\textbf{0.7779}  \textbf{(-)} &\textbf{0.6855}  \textbf{(-)}&\textbf{0.6762}  \textbf{(-)} &\textbf{0.8955}  \textbf{(-)}  \\
		&\textbf{M-RNN (SI)} &  \textbf{0.8530}  \textbf{(-)}  & \textbf{0.7783}  \textbf{(-)} &\textbf{0.6858}  \textbf{(-)}&\textbf{0.6759}  \textbf{(-)}&\textbf{0.8948}  \textbf{(-)}  \\
		\midrule
		\multirow{3}{*}{\textbf{RNN-based}}&\cite{Recent_RNN1} & 0.8381  (9.3\%)& 0.7558  (9.0\%)  &0.6505 (10.0\%) &0.6557 (6.0\%)&0.8802 (12.8\%)\\
		&\cite{Recent_RNN2}   & 0.8402  (8.1\%)&0.7551  (9.3\%)&0.6574 (8.2\%)  &0.6561 (5.8\%)&0.8748 (16.5\%)\\
		&\cite{Recent_RNN3}   & 0.8410  (7.6\%)&0.7593  (7.7\%)&0.6583 (8.0\%)  &0.6520  (7.0\%)&0.8826 (11.0\%)\\
		\midrule
		\multirow{2}{*}{\textbf{Interpolation}} & Spline &  0.8407 (7.8\%)& 0.7542  (9.6\%)&0.6477 (10.7\%)&0.6520 (7.0\%) &0.8731 (17.7\%)\\
		& Cubic   & 0.8397 (8.4\%)&0.7569 (8.6\%)&0.6468 (11.0\%)&0.6517  (7.0\%)&0.8643  (23.0\%)\\
		\midrule
		\multirow{3}{*}{\textbf{Imputation}} & MICE & 0.8377 (9.5\%)  &0.7571 (8.6\%)&0.6397 (12.7\%) &0.6509 (7.2\%)&0.8850  (9.1\%) \\
		&MissForest & 0.8368 (10.0\%)  & 0.7578 (8.3\%) &0.6740 (3.5\%)& 0.6587 (5.1\%)&0.8767 (15.2\%)\\		
		&EM   &  0.8312 (13.0\%)&0.7531 (10.0\%)& -& -&0.8794  (13.3\%) \\ 
		\midrule
		\multirow{3}{*}{\textbf{Others}}&Matrix Completion   &  0.8401 (8.1\%)&0.7551 (9.3\%)&0.6712 (4.3\%)& 0.6579 (5.3\%)&0.8865  (7.9\%) \\
		& Auto-encoder   & 0.8399 (8.2\%) &0.7488 (11.6\%)&0.6633 (6.6\%) &0.6574 (5.5\%) &0.8785 (14.0\%) \\
		& MCMC   & 0.8298 (13.7\%) &0.7512 (10.7\%)&0.6417 (12.2\%) &0.6512 (7.2\%)&0.8667 (21.6\%)  \\
		\bottomrule
	\end{tabular}
\end{table*}

Table \ref{tab:Prediction} shows the mean and percentage performance gain of M-RNN (MI) in comparison with the benchmarks on all the datasets.  M-RNN -- which we have already shown to achieve the best imputation accuracy -- also yields the best prediction accuracy. However, even in cases where the improvement in imputation accuracy is large and statistically significant, the  improvements in prediction accuracy are sometimes smaller and not  always statistically significant.  For instance, on the Deterioration dataset, the AUROC of M-RNN (MI) is 0.7779 (95\% CI: 0.7678-0.7868); the best benchmark is  \cite{Recent_RNN3} with AUROC of 0.7593 (95\% CI: 0.7478-0.7702). Similarly, on the UNOS-Heart dataset, the AUROC of M-RNN (MI) is 0.6855 (95\% CI: 0.6781-0.6913); the best benchmark is MissForest, with AUROC of 0.6740 (95\% CI: 0.6651-0.6817).

It should be noted that, by using mean square error as the loss function, we have deliberately optimized M-RNN for {\em imputation accuracy}.  If we want to optimize M-RNN for {\em prediction accuracy} we might do better by using a different loss function, such as cross-entropy.  In Table \ref{tab:Sup_Prediction} of the Appendix we report an experiment in which we have done precisely that; the short summary is that optimizing for prediction accuracy does in fact improve the predictive performance of M-RNN  but the improvement is marginal.

The Appendix also reports other  experiments that help further our understanding of the M-RNN algorithm.  Table \ref{tab:AddPrediction} demonstrates that using a different predictive model (random forest, logistic regression or Xgboost \cite{xgboost}, rather than a 1-layer RNN) for prediction after imputation leads to results similar to those obtained above.  Tables \ref{tab:DonorImputation} and \ref{tab:DonorPrediction} demonstrate that accounting for donor features in the UNOS datasets makes little difference.   

\subsection{Prediction accuracy with various missing rates}

\begin{figure}[t!]
	\center
	\includegraphics[width=0.7\textwidth]{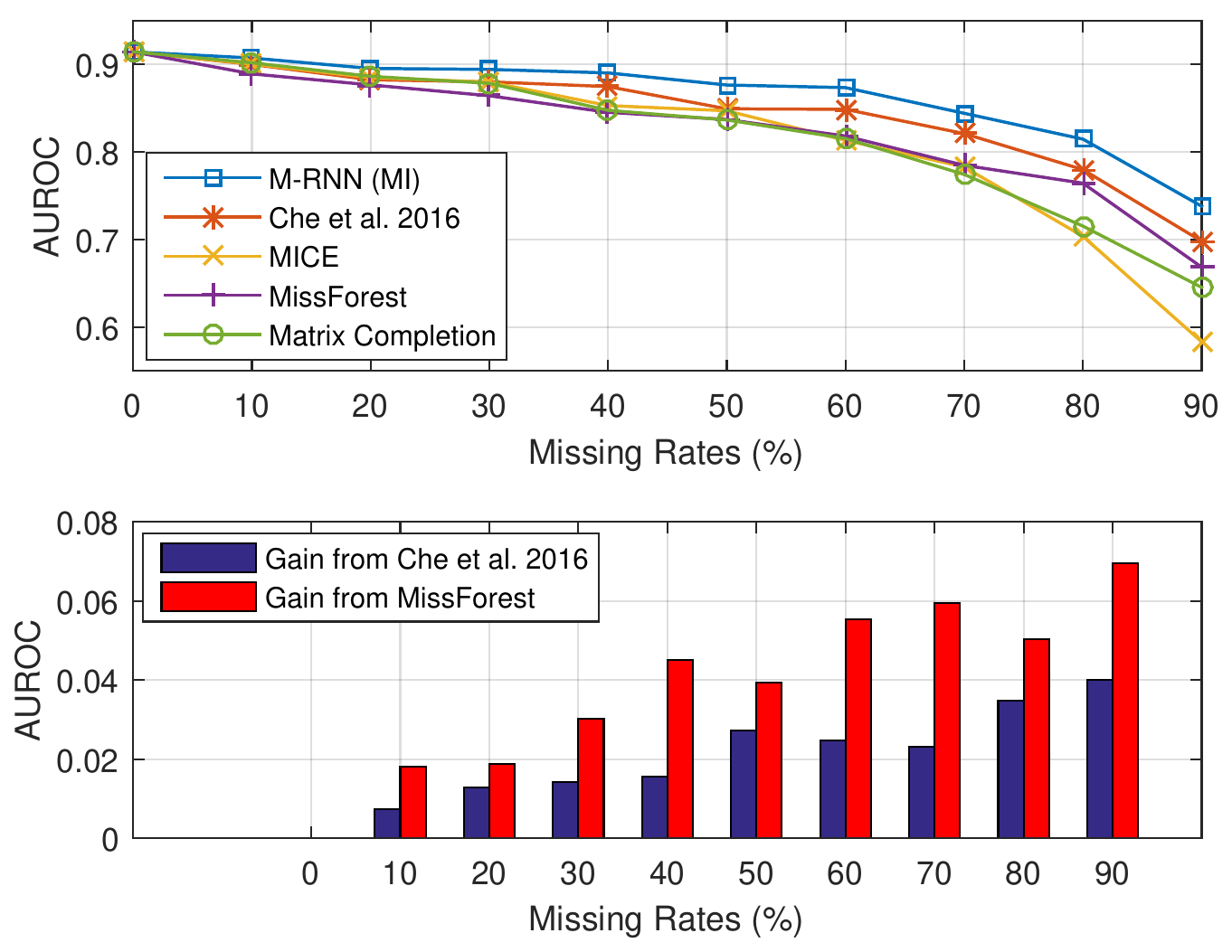}
	\caption{(a) The AUROC performance with various missing rates - upper line graph, (b) The AUROC gain from two most competitive benchmarks - lower bar graph}
	\label{fig:Biobank}
\end{figure}

As discussed above, we carried out  experiments with increased rates of missing data in order to understand the implications for the accuracy of imputation.  We also carried out  experiments with increased rates of missing data in order to understand the implications for the accuracy of prediction.  In this case, we begin with the Biobank dataset, which is complete, and randomly remove 10\% to 90\% of the measurements (with increments of 10\%) to create multiple datasets with different missing rates. (In each case we use 80\% of the data for training and 20\% for testing.)  As before, we use M-RNN and various benchmarks for imputing missing data and  a 1-layer RNN as the predictive model.  (In this setting we are predicting a clinical diagnosis of diabetes.)  In each setting, we conducted 10 trials, and report the performance  in terms of AUROC. 

Fig. \ref{fig:Biobank} (a) illustrates the impact (in terms of AUROC) of more and more missing data for M-RNN and various benchmarks. As Figure \ref{fig:Biobank} (a) shows, for M-RNN and all benchmarks, the prediction performance decreases as the amount of missing data increases.  However, as Fig. \ref{fig:Biobank} (b) shows, M-RNN continues to outperform the benchmarks; indeed, the performance gap between M-RNN and the benchmarks widens when more data is missing.  

\subsection{The importance of specific features}

\begin{figure}[t!]
	\centering
	\includegraphics[width=0.7\textwidth]{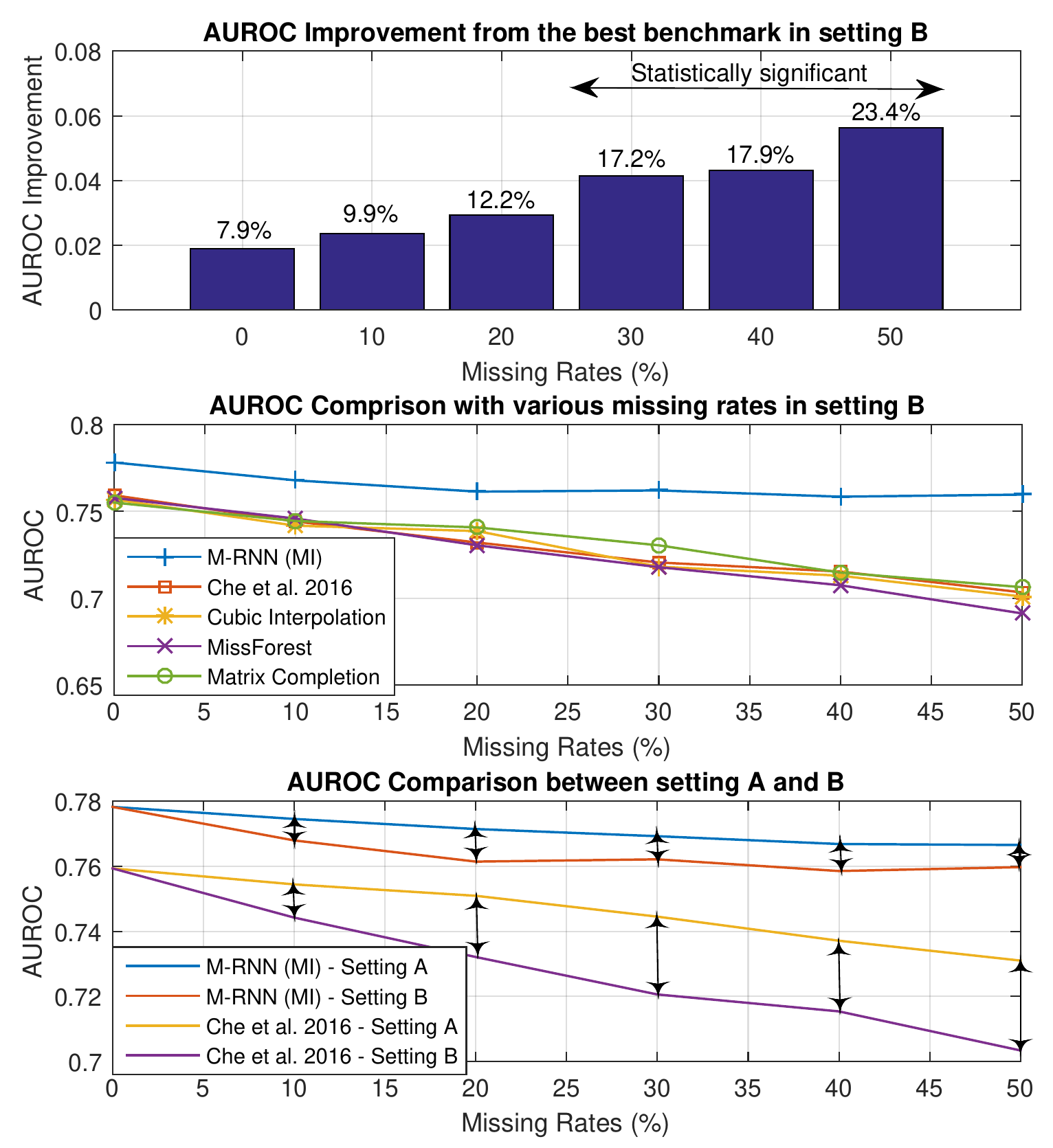}
	\caption{AUROC comparisons in Settings A and B using Deterioration dataset (a) AUROC improvement from the best benchmark (top) in Setting B, (b) AUROC comparisons with various missing rates in Setting B (middle), (c) AUROC comparison between Settings A and B (bottom)}
	\label{fig:correlate}
\end{figure} 

To this point, we have treated all missing data as equally important and given the same weight to all errors.  However, this is not always the right thing to do.  In particular, it is clear that not all missing data is equally important for prediction.    To understand the importance of missing data for purposes of prediction we conduct two experiments in parallel.  For the first experiment (which we call Setting A: Purely Random Removal), we construct 5 sub-samples of the Deterioration dataset by randomly removing 10\%, 20\%, 30\%, 40\%, 50\% of the measurements for randomly chosen features. For the second experiment (which we call Setting B: Correlated Random Removal)  we first identify the four features that are most highly correlated with the label (in the Deterioration dataset the label means either Admission to the ICU or Discharge from Hospital); these four features are: O2 Device, Heart rates, Respiratory rate and Urea Nitrogen.  We then construct 5 sub-samples of the Deterioration dataset by removing 10\%, 20\%, 30\%, 40\%, 50\% of the measurements for {\em these  specific features}.  In both cases we repeat the exercise 10 times and report average results.   We then compare the prediction performance of M-RNN (MI) with the best benchmarks; the results are shown visually in Fig. \ref{fig:correlate}.

Because the purpose of this exercise is to see the impact of missingness of more important features on prediction, we focus most of our attention on Setting B. Fig. \ref{fig:correlate}(b) shows the AUROC for M-RNN and benchmarks for various rates of missing data in Setting B.  As can be seen, M-RNN outperforms the best benchmarks for every sub-sample and the improvement in performance is greater for the sub-samples for which more data is missing.  Fig. \ref{fig:correlate}(a) shows the percentage improvement of M-RNN against the benchmarks (still in Setting B).  The improvement in performance is statistically significant (p-value $<$ 0.05) when 30\% or more of the measurements (of the most important features) are missing.  Fig. \ref{fig:correlate}(c) shows the prediction performance of M-RNN and the best benchmark in Settings A and B.  As can be seen, the prediction performances of both M-RNN and the benchmark are worse when the more important data is missing -- but the prediction performance of M-RNN is much less sensitive to the amount of data that is missing and to which data is missing.  

\subsection{Congeniality of the model}\label{sect:congenial}
As \cite{congeniality_meng} has emphasized, an extremely desirable aspect of any imputation method is that it produce imputed values in a manner that is consistent and preserves the original relationships between features and labels;  \cite{congeniality_meng}  refers to this as {\em congeniality}.  Congeniality of an imputation model can be evaluated with respect to a particular model of the feature-label relationships by computing the model parameters for the true complete data and the imputed data and measuring the difference between parameters according to some specified metric.  Of course no imputation method can be expected to be perfectly congenial, but we argue that our method is more congenial -- i.e. better preserves the relationships between features and labels -- than benchmarks.  To see this, we exploit the fact that the Biobank dataset is complete so that it is possible to delete some of the data, impute the deleted data, and compare the relationship between the actual (original) data and labels with the relationship between the imputed data and labels.  

In our particular experiment, we delete 20\% of the data and impute the missing data using our M-RNN and the 4 best benchmarks (the method of \cite{Recent_RNN3}, Cubic Interpolation, MissForest and Matrix Completion).  As a model of the feature-label relationship, we use a logistic regression.  As a metric of the difference between the logistic regression parameters $\mathbf{w}$ for the actual data and $\hat{\mathbf{w}}$ for the imputed data (which  can be interpreted as a measure of the uncongeniality of the imputation) we report both the mean bias $\Vert \mathbf{w} - \hat{\mathbf{w}} \Vert_1$ and  the root mean square error error $\Vert \mathbf{w} - \hat{\mathbf{w}} \Vert_2$.  

\begin{table}[h!]
	\caption{Congeniality of imputation models}
	\label{tab:congeniality}
	\centering
	\small
	\begin{tabular}{c|c|c}
		\toprule
		\multirow{2}{*}{\textbf{Algorithm}} & \textbf{Mean Bias} & \textbf{Root Mean Square Error} \\
		& $(||\mathbf{w} - \hat{\mathbf{w}}||_1)$ & $(||\mathbf{w} - \hat{\mathbf{w}}||_2)$\\
		\midrule
		\textbf{M-RNN (MI)} & 0.0814 $\pm$ 0.0098 & 0.1229 $\pm$0.0151  \\
		\midrule
		\cite{Recent_RNN3}   & 0.1097 $\pm$ 0.0104 & 0.1649 $\pm$ 0.0212 \\
		\midrule
		Cubic Interpolation & 0.1169 $\pm$ 0.01075  & 0.1816 $\pm$ 0.0201 \\
		\midrule
		MissForest & 0.0842 $\pm$0.0103 &0.1312 $\pm$ 0.0139\\	
		\midrule
		Matrix Completion   & 0.1001 $\pm$ 0.0125 & 0.1551 $\pm$ 0.0230 \\
		\bottomrule
	\end{tabular}
\end{table}

As can be seen in Table \ref{tab:congeniality}, in comparison with the 4 best benchmarks, M-RNN achieves both smaller mean bias and small root mean square error between the original and imputed representations of feature-label relationship.  (With the exception of MissForest, all the performance improvements of M-RNN are statistically significant at the 95\% level.)  Thus our method is more congenial (to the logistic regression model) than the benchmarks.

\section{Conclusion}
The problem of reconstructing/estimating missing data is ubiquitous in many settings -- especially  in longitudinal medical datasets -- and is of enormous importance for many reasons, including statistical analysis, diagnosis, prognosis and treatment.  In this paper we have presented a new  method, based on a novel deep learning architecture, for reconstructing/estimating missing data that exploits both the correlation within data streams and the correlation across data streams. We have demonstrated on the basis of a variety of real-world medical datasets that our method makes large and statistically significant improvements in comparison with state-of-the-art benchmarks.

    \section*{Acknowledgment}
    We are grateful for comments from Katrina Poppe (University of Auckland) and Angela Wood (University of Cambridge). This work was supported by the Office of Naval Research (ONR) and the NSF (Grant number: ECCS1462245).


    \newpage
\section*{Appendix}

\subsection*{Predictions with Alternative Predictive Models }

\begin{table*}[h!]
	\caption{Performance comparison for patient state prediction using Deterioration dataset}
	\label{tab:AddPrediction}
	\centering
	\small
	\begin{tabular}{c|c|c|c|c|c}
		\toprule
		\multirow{2}{*}{\textbf{Category}} & \multirow{2}{*}{\textbf{Algorithms}} &\multicolumn{4}{c}{\textbf{AUROC (\% Gain of M-RNN (Multiple Imputations))}}  \\
		\cmidrule{3-6}
		& & RNN & Random Forest & Logistic Regression & Xgboost \\
		\midrule
		\multirow{2}{*}{\textbf{M-RNN}}&\textbf{M-RNN (MI)} &  \textbf{0.7779}  \textbf{(-)}  &\textbf{0.7568}  \textbf{(-)} &\textbf{0.7657}  \textbf{(-)}&\textbf{0.7619}  \textbf{(-)} \\
		&\textbf{M-RNN (SI)} &  \textbf{0.7783}  \textbf{(-)}  & \textbf{0.7577}  \textbf{(-)} &\textbf{0.7652}  \textbf{(-)}&\textbf{0.7599}  \textbf{(-)} \\
		\midrule
		\multirow{3}{*}{\textbf{RNN-based}} &\cite{Recent_RNN1} &  0.7558  (9.0\%)  &-  &- &-\\
		&\cite{Recent_RNN2}   & 0.7551  (9.3\%)&-  &- &-\\
		&\cite{Recent_RNN3}   &0.7593  (7.7\%)&-  &- &-\\
		\midrule
		\multirow{3}{*}{\textbf{Interpolation}}&Spline    &  0.7542 (9.6\%)& 0.7396  (6.6\%)&0.7504 (6.1\%)&0.7442 (6.9\%) \\
		& Cubic   & 0.7569 (8.6\%)&0.7348 (8.3\%)&0.7475 (7.2\%)&0.7406  (8.2\%)\\
		\midrule
		\multirow{3}{*}{\textbf{Imputation}}& MICE & 0.7571 (8.6\%)  &0.7315 (9.4\%)&0.7422 (9.1\%) &0.7358 (9.9\%) \\
		& MissForest & 0.7578 (8.3\%)  & 0.7331 (8.9\%) &0.7459 (7.8\%)& 0.7403 (8.3\%)\\		
		& EM    &  0.7531 (10.0\%)&0.7252 (11.5\%)& 0.7355 (11.4\%)& 0.7286 (12.3\%) \\ \hline
		\multirow{3}{*}{\textbf{Others}}& Matrix Completion   &  0.7551 (9.3\%)&0.7272 (10.9\%)&0.7454 (8.0\%)& 0.7385 (8.9\%) \\
		&Auto-encoder   & 0.7488 (11.6\%) &0.7256 (11.4\%)&0.7343 (11.8\%) &0.7279 (12.5\%)  \\
		&MCMC   & 0.7512 (10.7\%) &0.7238 (11.9\%)&0.7325 (12.4\%) &0.7270 (12.8\%)  \\
		\bottomrule
	\end{tabular}
\end{table*}

Table \ref{tab:Prediction} compares the implied predictive performance of M-RNN with the benchmarks when imputation is followed by prediction using a particularly simple predictive model.  We now compare the implied predictive performance of  M-RNN with the benchmarks when imputation is followed by prediction other predictive models: in addition to the 1-layer RNN model used already, we use Random Forest \cite{rforest}, Logistic Regression and Xgboost \cite{xgboost}. (For implementations, we use the randomForest package, the glm package and the xgboost package, all in R.)  We restrict our attention to the Deterioration dataset and continue to use AUROC as the performance metric.  Table \ref{tab:AddPrediction} shows the mean and performance gain (\%) (in terms of AUROC) of M-RNN in comparison with the benchmarks with various predictive models.  

As can be seen in  Table \ref{tab:AddPrediction}, the different predictive models do not yield much different prediction accuracy, and no one predictive model seems to consistently lead to more or less accurate predictions. (Because the RNN imputation methods \cite{Recent_RNN1,Recent_RNN2,Recent_RNN3} can only be combined with an RNN predictive model, there are no RF, LR, XG results for these methods.) Note that  prediction accuracy is {\em not} perfectly correlated with  imputation accuracy, but the differences are not statistically significant.  As before, M-RNN leads to the most accurate predictions when followed by each of the various predictive models; again, not all the differences are statistically significant. 	

\subsection*{Effects of Donor Features in UNOS Datasets}
The UNOS organ transplant dataset records features of both the recipient and the donor.  In the main text we ignored donor features; here we expand the analysis to include these features, and compare the results obtained  using only recipient features  with the results obtained using both recipient and donor features. Table \ref{tab:DonorImputation} presents the comparison for imputations and Table \ref{tab:DonorPrediction} presents the comparisons for prediction. As noted in the main text, the differences are not large, presumably because the donor information is completely static and the relevant aspects are captured implicitly in the recipient data.
  
\begin{table*}[t!]
	\caption{Performance comparison for missing value estimation with/without donor features in UNOS dataset }
	\label{tab:DonorImputation}
	\centering
	\small
	\begin{tabular}{c|c|c|c|c|c}
		\toprule
		\multirow{2}{*}{\textbf{Category}}&\multirow{2}{*}{\textbf{Algorithms}}&\multicolumn{4}{c}{\textbf{Mean RMSE (\% Gain of M-RNN (Multiple Imputations))}}  \\
		\cmidrule{3-6}
		& & Heart (with) & Heart (w/o) & Lung (with) & Lung (w/o) \\
		\midrule
		\multirow{2}{*}{\textbf{M-RNN}}&\textbf{M-RNN (MI)} &  \textbf{0.0451}  \textbf{(-)}  &\textbf{0.0479}  \textbf{(-)}&\textbf{0.0579}  \textbf{(-)}&\textbf{0.0606}  \textbf{(-)} \\
		&\textbf{M-RNN (SI)} &  \textbf{0.0453}  \textbf{(-)}  &\textbf{0.0477}  \textbf{(-)}&\textbf{0.0583}  \textbf{(-)}&\textbf{0.0609}  \textbf{(-)} \\
		\midrule
		\multirow{3}{*}{\textbf{RNN-based}}&\cite{Recent_RNN1} & 0.1352 (66.6\%)&0.1352 (64.6\%) & 0.1343 (56.9\%) & 0.1343 (54.9\%) \\
		&\cite{Recent_RNN2} &0.1179 (61.7\%) &0.1179 (59.4\%)& 0.1264 (54.2\%) & 0.1264 (52.1\%)\\
		&\cite{Recent_RNN3} &0.1057 (57.3\%) &0.1057 (54.7\%)&  0.1172 (50.6\%) & 0.1172 (48.3\%) \\
		\midrule
		\multirow{2}{*}{\textbf{Interpolation}} & Spline    &0.1102  (59.1\%) &0.1102  (56.5\%)& 0.1199 (51.7\%) & 0.1199 (49.5\%)\\
		& Cubic  & 0.1072 (57.9\%)&0.1072 (55.3\%)&  0.1177 (50.8\%) & 0.1177 (48.5\%)\\
		\midrule
		\multirow{2}{*}{\textbf{Imputation}} & MICE &  0.1067 (57.7\%) &0.1147  (58.2\%)& 0.1015 (43.0\%) & 0.1151  (47.4\%)\\
		&MissForest & 0.0465 (3.0\%)& 0.0489  (2.0\%)& 0.0627 (7.7\%) & 0.0652  (7.1\%) \\ 
		\midrule
		\multirow{3}{*}{\textbf{Others}} & Matrix Completion   & 0.0767 (41.2\%) &0.0974  (50.8\%)& 0.0819 (29.3\%)& 0.0942 (35.7\%)  \\
		&Auto-encoder   & 0.0544 (17.1\%) &0.0589 (18.7\%) & 0.0668 (13.3\%) & 0.0712 (14.9\%) \\		
		& MCMC   & 0.0934 (51.7\%) &0.1091 (56.1\%) & 0.1017 (43.1\%) & 0.1124 (46.1\%) \\				
		\bottomrule
	\end{tabular}
\end{table*}

As can be seen in  Tables \ref{tab:DonorImputation} and \ref{tab:DonorPrediction}, the effects of donor features on imputation and prediction accuracy are not large.  The RMSE of M-RNN (MI) improves from from 0.0479 to 0.0451  (5.8\%) on the UNOS-Heart dataset and 0.0606 to 0.0579   (4.5\%) on the UNOS-Lung dataset.  In terms of prediction, AUROC of M-RNN (MI) improves from from 0.6855 to 0.7153  (9.5\%) on the  UNOS-Heart dataset and from 0.6762  to 0.6883 (3.7\%) on  UNOS-Lung dataset.   
\begin{table*}[t!]
	\caption{Performance comparison for label prediction (AUROC) with/without donor features in UNOS dataset}
	\label{tab:DonorPrediction}
	\centering
	\small
	\begin{tabular}{c|c|c|c|c|c}
		\toprule
		\multirow{2}{*}{\textbf{Category}} & \multirow{2}{*}{\textbf{Algorithm}}&\multicolumn{4}{c}{\textbf{AUROC (\% Gain of M-RNN (Multiple Imputations))}}  \\
		\cmidrule{3-6}
		& & Heart (with) & Heart (w/o) & Lung (with) & Lung (w/o) \\
		\midrule
		\multirow{2}{*}{\textbf{M-RNN}}&\textbf{M-RNN (MI)} &  \textbf{0.7153}  \textbf{(-)}  &\textbf{0.6855}  \textbf{(-)} & \textbf{0.6883}  \textbf{(-)} & \textbf{0.6762}  \textbf{(-)} \\
		&\textbf{M-RNN (SI)} &  \textbf{0.7141}  \textbf{(-)}  & \textbf{0.6858}  \textbf{(-)} & \textbf{0.6886}  \textbf{(-)} & \textbf{0.6759}  \textbf{(-)}  \\
		\midrule
		\multirow{3}{*}{\textbf{RNN-based}}&\cite{Recent_RNN1} &  0.6830 (10.2\%)& 0.6505  (10.0\%)  & 0.6575 (9.0\%) & 0.6557(6.0\%) \\
		&\cite{Recent_RNN2}   & 0.6892 (8.4\%)& 0.6574 (8.2\%)& 0.6651 (6.9\%) & 0.6561(5.8\%) \\
		&\cite{Recent_RNN3}   &  0.6962 (6.3\%)& 0.6583 (8.0\%)& 0.6728 (4.7\%) &0.6520 (7.0\%) \\
		\midrule
		\multirow{2}{*}{\textbf{Interpolation}}& Spline & 0.6817 (10.6\%)& 0.6477(10.7\%)&0.6651 (6.9\%)& 0.6520 (7.0\%) \\
		& Cubic & 0.6788 (11.4\%) & 0.6468 (11.0\%) &0.6629 (7.5\%) &0.6517 (7.0\%) \\
		\midrule
		\multirow{2}{*}{\textbf{Imputation}}& MICE & 0.6785 (11.4\%)  & 0.6397 (12.7\%) & 0.6567 (9.2\%) & 0.6509 (7.2\%) \\
		& MissForest & 0.6981 (5.7\%)  &0.6740 (1.7\%) &0.6754 (4.0\%) &0.6587 (5.1\%) \\ 
		\midrule
		\multirow{3}{*}{\textbf{Others}} & Matrix Completion   & 0.6969 (6.1\%) &0.6712 (2.2\%) &0.6745 (4.2\%) &0.6579 (5.3\%) \\
		& Auto-encoder  &0.6966 (6.2\%) &0.6633 (6.6\%) &0.6741 (4.4\%) &0.6574 (5.5\%)  \\
		& MCMC  &0.6772 (11.8\%) &0.6417 (12.2\%) &0.6717 (5.1\%) &0.6512 (7.2\%)  \\
		\bottomrule
	\end{tabular}
\end{table*}

\subsection*{Prediction-oriented M-RNN}

\begin{table*}[h!]
	\caption{Performance comparison for prediction oriented M-RNN}
	\label{tab:Sup_Prediction}
	\centering
	\small
	\begin{tabular}{c|c|c|c|c|c|c}
		\toprule
		\multirow{2}{*}{\textbf{Algorithms}} &\multicolumn{5}{c}{\textbf{AUROC (\% Gain of M-RNN (Multiple Imputations))}}  \\
		\cmidrule{2-6}
		 & MIMIC-III & Deterioration & UNOS-Heart & UNOS-Lung & Biobank \\
		\midrule
		\textbf{M-RNN: Predict} &  \textbf{0.8578}  \textbf{(-)}  & \textbf{0.7813}  \textbf{(-)} &\textbf{0.6897}  \textbf{(-)}&\textbf{0.6802}  \textbf{(-)} &\textbf{0.8987}  \textbf{(-)}\\
		\textbf{M-RNN: Original} &  \textbf{0.8531}  \textbf{(3.2\%)}  &\textbf{0.7779}  \textbf{(1.5\%)} &\textbf{0.6855}  \textbf{(1.3\%)}&\textbf{0.6762}  \textbf{(1.2\%)}&\textbf{0.8955}  \textbf{(3.1\%)} \\
		\midrule
		\cite{Recent_RNN1} & 0.8381  (12.2\%)& 0.7558  (10.4\%)  &0.6505 (11.2\%) &0.6557 (7.1\%)&0.8802 (15.4\%)\\
		\cite{Recent_RNN2}   & 0.8402  (11.0\%)&0.7551  (10.7\%)&0.6574 (9.4\%)  &0.6561 (7.0\%) &0.8748 (19.1\%)\\
		\cite{Recent_RNN3}   & 0.8410  (10.6\%)&0.7593  (9.1\%)&0.6583 (9.2\%)  &0.6520  (8.1\%)&0.8826 (13.7\%)\\
		\bottomrule
	\end{tabular}
\end{table*}

Finally, we report the results of an experiment in which we optimized M-RNN for prediction accuracy rather than imputation accuracy.  To do this, we trained M-RNN to minimize the cross-entropy $\mathcal{L} = \frac{1}{N} \sum_{n=1}^N [y_n \log \hat{y}_n + (1 - y_n) \log (1 - \hat{y}_n )]$, rather than the mean square error. (We continue to evaluate performance in terms of AUROC.)  As can be seen from Table \ref{tab:Sup_Prediction}, doing this does improve the predictions of M-RNN, but the improvement is marginal and not statistically significant.  (However, doing this does have the advantage that it creates an end-to-end prediction algorithm that does not require any  preprocessing or imputation steps.)

\subsection*{Implementations}
For some of these algorithms, we are able to use off-the-shelf implementations.  For Spline and Cubic Interpolation, we use the interp1 package in MATLAB; for MICE we use the mice package in R; for MissForest we use the MissForest package in R; for EM we use the Amelia package in R; for matrix completion we use the softImpute package in R.
    
    
    %


\begin{thebibliography}{10}
	\providecommand{\url}[1]{#1}
	\csname url@samestyle\endcsname
	\providecommand{\newblock}{\relax}
	\providecommand{\bibinfo}[2]{#2}
	\providecommand{\BIBentrySTDinterwordspacing}{\spaceskip=0pt\relax}
	\providecommand{\BIBentryALTinterwordstretchfactor}{4}
	\providecommand{\BIBentryALTinterwordspacing}{\spaceskip=\fontdimen2\font plus
		\BIBentryALTinterwordstretchfactor\fontdimen3\font minus
		\fontdimen4\font\relax}
	\providecommand{\BIBforeignlanguage}[2]{{%
			\expandafter\ifx\csname l@#1\endcsname\relax
			\typeout{** WARNING: IEEEtran.bst: No hyphenation pattern has been}%
			\typeout{** loaded for the language `#1'. Using the pattern for}%
			\typeout{** the default language instead.}%
			\else
			\language=\csname l@#1\endcsname
			\fi
			#2}}
	\providecommand{\BIBdecl}{\relax}
	\BIBdecl
	
	\bibitem{interpolation}
	D.~M. Kreindler and C.~J. Lumsden, ``The effects of the irregular sample and
	missing data in time series analysis,'' \emph{Nonlinear Dynamical Systems
		Analysis for the Behavioral Sciences Using Real Data}, p. 135, 2012.
	
	\bibitem{wavelet}
	D.~Mondal and D.~B. Percival, ``Wavelet variance analysis for gappy time
	series,'' \emph{Annals of the Institute of Statistical Mathematics}, vol.~62,
	no.~5, pp. 943--966, 2010.
	
	\bibitem{Rubin}
	D.~B. Rubin, \emph{Multiple imputation for nonresponse in surveys}.\hskip 1em
	plus 0.5em minus 0.4em\relax John Wiley \& Sons, 2004, vol.~81.
	
	\bibitem{EM}
	P.~J. Garc{\'\i}a-Laencina, J.-L. Sancho-G{\'o}mez, and A.~R. Figueiras-Vidal,
	``Pattern classification with missing data: a review,'' \emph{Neural
		Computing and Applications}, vol.~19, no.~2, pp. 263--282, 2010.
	
	\bibitem{MICE}
	I.~R. White, P.~Royston, and A.~M. Wood, ``Multiple imputation using chained
	equations: issues and guidance for practice,'' \emph{Statistics in medicine},
	vol.~30, no.~4, pp. 377--399, 2011.
	
	\bibitem{missforest}
	D.~J. Stekhoven and P.~B{\"u}hlmann, ``Missforest—non-parametric missing
	value imputation for mixed-type data,'' \emph{Bioinformatics}, vol.~28,
	no.~1, pp. 112--118, 2011.
	
	\bibitem{Mat-0}
	R.~Mazumder, T.~Hastie, and R.~Tibshirani, ``Spectral regularization algorithms
	for learning large incomplete matrices,'' \emph{Journal of machine learning
		research}, vol.~11, no. Aug, pp. 2287--2322, 2010.
	
	\bibitem{Mat-1}
	H.-F. Yu, H.~Rao, and I.~S. Dhillon, ``Temporal regularized matrix
	factorization for high-dimensional time series prediction,'' \emph{Proc
		NIPS}, 2016.
	
	\bibitem{Mat-2}
	T.~Schnabel, A.~Swaminatan, A.~Singh, N.~Chandak, and T.~Joachims,
	``Recommendations as treatments: debiasing learning and evolution,''
	\emph{Proc ICML}, 2016.
	
	\bibitem{Mat-3}
	R.~Mazumder, T.~Hastie, and R.~Tibshirani, ``Spectral regularization algorithms
	for learning large incomplete matrices,'' \emph{Journal of machine learning
		research}, vol.~11, no. Aug, pp. 2287--2322, 2010.
	
	\bibitem{Missing_Book}
	D.~M. Kreindler and C.~J. Lumsden, ``The effects of the irregular sample and
	missing data in time series analysis,'' \emph{Nonlinear Dynamical Systems
		Analysis for the Behavioral Sciences Using Real Data}, p. 135, 2012.
	
	\bibitem{ICML2017-Alaa}
	A.~M. Alaa, S.~Hu, and M.~van~der Schaar, ``Learning from clinical judgments:
	Semi-markov-modulated marked hawkes processes for risk prognosis,'' in
	\emph{Proceedings of the 34th international conference on machine learning
		(ICML-17)}, 2017.
	
	\bibitem{BiRNN}
	A.~Graves and J.~Schmidhuber, ``Framewise phoneme classification with
	bidirectional lstm and other neural network architectures,'' \emph{Neural
		Networks}, vol.~18, no.~5, pp. 602--610, 2005.
	
	\bibitem{singleimpute}
	A.~R.~T. Donders, G.~J. van~der Heijden, T.~Stijnen, and K.~G. Moons, ``A
	gentle introduction to imputation of missing values,'' \emph{Journal of
		clinical epidemiology}, vol.~59, no.~10, pp. 1087--1091, 2006.
	
	\bibitem{Multipleimpute1}
	P.~A. Patrician, ``Multiple imputation for missing data,'' \emph{Research in
		Nursing \& Health}, vol.~25, no.~1, pp. 76--84, 2002.
	
	\bibitem{reason_of_imputation1}
	S.~Burgess, I.~R. White, M.~Resche-Rigon, and A.~M. Wood, ``Combining multiple
	imputation and meta-analysis with individual participant data,''
	\emph{Statistics in medicine}, vol.~32, no.~26, pp. 4499--4514, 2013.
	
	\bibitem{Medimpute2}
	A.~Mackinnon, ``The use and reporting of multiple imputation in medical
	research--a review,'' \emph{Journal of internal medicine}, vol. 268, no.~6,
	pp. 586--593, 2010.
	
	\bibitem{Medimpute3}
	J.~A. Sterne, I.~R. White, J.~B. Carlin, M.~Spratt, P.~Royston, M.~G. Kenward,
	A.~M. Wood, and J.~R. Carpenter, ``Multiple imputation for missing data in
	epidemiological and clinical research: potential and pitfalls,'' \emph{Bmj},
	vol. 338, p. b2393, 2009.
	
	\bibitem{Dropout}
	N.~Srivastava, G.~E. Hinton, A.~Krizhevsky, I.~Sutskever, and R.~Salakhutdinov,
	``Dropout: a simple way to prevent neural networks from overfitting.''
	\emph{Journal of Machine Learning Research}, vol.~15, no.~1, pp. 1929--1958,
	2014.
	
	\bibitem{MIMIC}
	A.~E. Johnson, T.~J. Pollard, L.~Shen, L.-w.~H. Lehman, M.~Feng, M.~Ghassemi,
	B.~Moody, P.~Szolovits, L.~A. Celi, and R.~G. Mark, ``Mimic-iii, a freely
	accessible critical care database,'' \emph{Scientific data}, vol.~3, 2016.
	
	\bibitem{TBME}
	A.~M. Alaa, J.~Yoon, S.~Hu, and M.~van~der Schaar, ``Personalized risk scoring
	for critical care prognosis using mixtures of gaussian processes,''
	\emph{IEEE Transactions on Biomedical Engineering}, 2017.
	
	\bibitem{Biobank}
	L.~J. Palmer, ``Uk biobank: bank on it,'' \emph{The Lancet}, vol. 369, no.
	9578, pp. 1980--1982, 2007.
	
	\bibitem{Recent_RNN1}
	E.~Choi, M.~T. Bahadori, and J.~Sun, ``Doctor ai: Predicting clinical events
	via recurrent neural networks,'' \emph{arXiv preprint arXiv:1511.05942},
	2015.
	
	\bibitem{Recent_RNN2}
	Z.~C. Lipton, D.~C. Kale, and R.~Wetzel, ``Directly modeling missing data in
	sequences with rnns: Improved classification of clinical time series,''
	\emph{arXiv preprint arXiv:1606.04130}, 2016.
	
	\bibitem{Recent_RNN3}
	Z.~Che, S.~Purushotham, K.~Cho, D.~Sontag, and Y.~Liu, ``Recurrent neural
	networks for multivariate time series with missing values,'' \emph{arXiv
		preprint arXiv:1606.01865}, 2016.
	
	\bibitem{reason_of_imputation2}
	Y.~Deng, C.~Chang, M.~S. Ido, and Q.~Long, ``Multiple imputation for general
	missing data patterns in the presence of high-dimensional data,''
	\emph{Scientific reports}, vol.~6, p. 21689, 2016.
	
	\bibitem{congeniality_meng}
	X.-L. Meng, ``Multiple-imputation inferences with uncongenial sources of
	input,'' \emph{Statistical Science}, pp. 538--558, 1994.
	
	\bibitem{Early_RNN1}
	F.~Gingras and Y.~Bengio, ``Recurrent neural networks for missing or
	asynchronous data,'' in \emph{Proc NIPS}, vol.~8, 1996.
	
	\bibitem{Early_RNN2}
	V.~Tresp and T.~Briegel, ``A solution for missing data in recurrent neural
	networks with an application to blood glucose prediction,'' \emph{Advances in
		Neural Information Processing Systems}, pp. 971--977, 1998.
	
	\bibitem{Early_RNN3}
	S.~Parveen and P.~Green, ``Speech recognition with missing data using recurrent
	neural nets,'' in \emph{Advances in Neural Information Processing Systems},
	2002, pp. 1189--1195.
	
	\bibitem{Recent_RNN5}
	H.-G. Kim, G.-J. Jang, H.-J. Choi, M.~Kim, Y.-W. Kim, and J.~Choi, ``Recurrent
	neural networks with missing information imputation for medical examination
	data prediction,'' in \emph{Big Data and Smart Computing (BigComp), 2017 IEEE
		International Conference on}.\hskip 1em plus 0.5em minus 0.4em\relax IEEE,
	2017, pp. 317--323.
	
	\bibitem{autoencoder}
	L.~Gondara and K.~Wang, ``Multiple imputation using deep denoising
	autoencoders,'' \emph{arXiv preprint arXiv:1705.02737}, 2017.
	
	\bibitem{mcmc}
	D.~Schunk, ``A markov chain monte carlo algorithm for multiple imputation in
	large surveys,'' \emph{AStA Advances in Statistical Analysis}, vol.~92,
	no.~1, pp. 101--114, 2008.
	
	\bibitem{xgboost}
	T.~Chen and C.~Guestrin, ``Xgboost: A scalable tree boosting system,'' in
	\emph{Proceedings of the 22nd acm sigkdd international conference on
		knowledge discovery and data mining}.\hskip 1em plus 0.5em minus 0.4em\relax
	ACM, 2016, pp. 785--794.
	
	\bibitem{rforest}
	L.~Breiman, ``Random forests,'' \emph{Machine learning}, vol.~45, no.~1, pp.
	5--32, 2001.
	
\end{thebibliography}
\end{document}